%% file: main.tex
\documentclass[10pt,twocolumn,letterpaper]{article}

\usepackage[pagenumbers]{cvpr} % To force page numbers, e.g. for an arXiv version

\input{preamble}

\definecolor{cvprblue}{rgb}{0.21,0.49,0.74}
\usepackage[pagebackref,breaklinks,colorlinks,allcolors=cvprblue]{hyperref}
\usepackage{multirow}
\usepackage{soul}
\usepackage{url}
\usepackage{color}
\usepackage{balance} 
\usepackage{xspace}
\usepackage{tabularx}
\usepackage{colortbl}
\usepackage{enumitem}
\usepackage{wasysym}
\usepackage{pifont}
\usepackage{xcolor}

\definecolor{bblue}{rgb}{0,150,230}
\definecolor{mygray}{gray}{.9}
\definecolor{lightgray}{gray}{.96}
\definecolor{myy}{RGB}{126,95,0}
\definecolor{ggray}{RGB}{127,127,127}
\definecolor{mygreen}{RGB}{93,173,85}
\definecolor{myred}{RGB}{240,16,89}
\definecolor{myblue}{RGB}{0,114,188}
\definecolor{darkgreen}{rgb}{0.0, 0.5, 0.0}
\definecolor{demphcolor}{RGB}{100,100,100}
\definecolor{sh_blue}{rgb}{0,0.60,0.93}
\definecolor{sh_red}{rgb}{0.8627, 0.3098, 0.3176}
\definecolor{lightpink}{rgb}{0.918, 0.761, 0.761}
\definecolor{lightblue}{rgb}{0.671, 0.773, 0.863}
\definecolor{customlightgreen}{rgb}{0.8196, 0.8784, 0.7098}
\definecolor{lightpeach}{rgb}{1.0, 0.882, 0.788}
\definecolor{customblue}{rgb}{0.180, 0.400, 0.522}
\definecolor{lightcyan}{rgb}{0.8196, 0.9725, 0.9804}

\newcommand{\yeying}[1]{\textcolor{black}{#1}}

\usepackage[normalem]{ulem}

\def\paperID{4216} % *** Enter the Paper ID here
\def\confName{CVPR}
\def\confYear{2025}

\title{
\raisebox{-0.1cm}{\includegraphics[width=1cm]{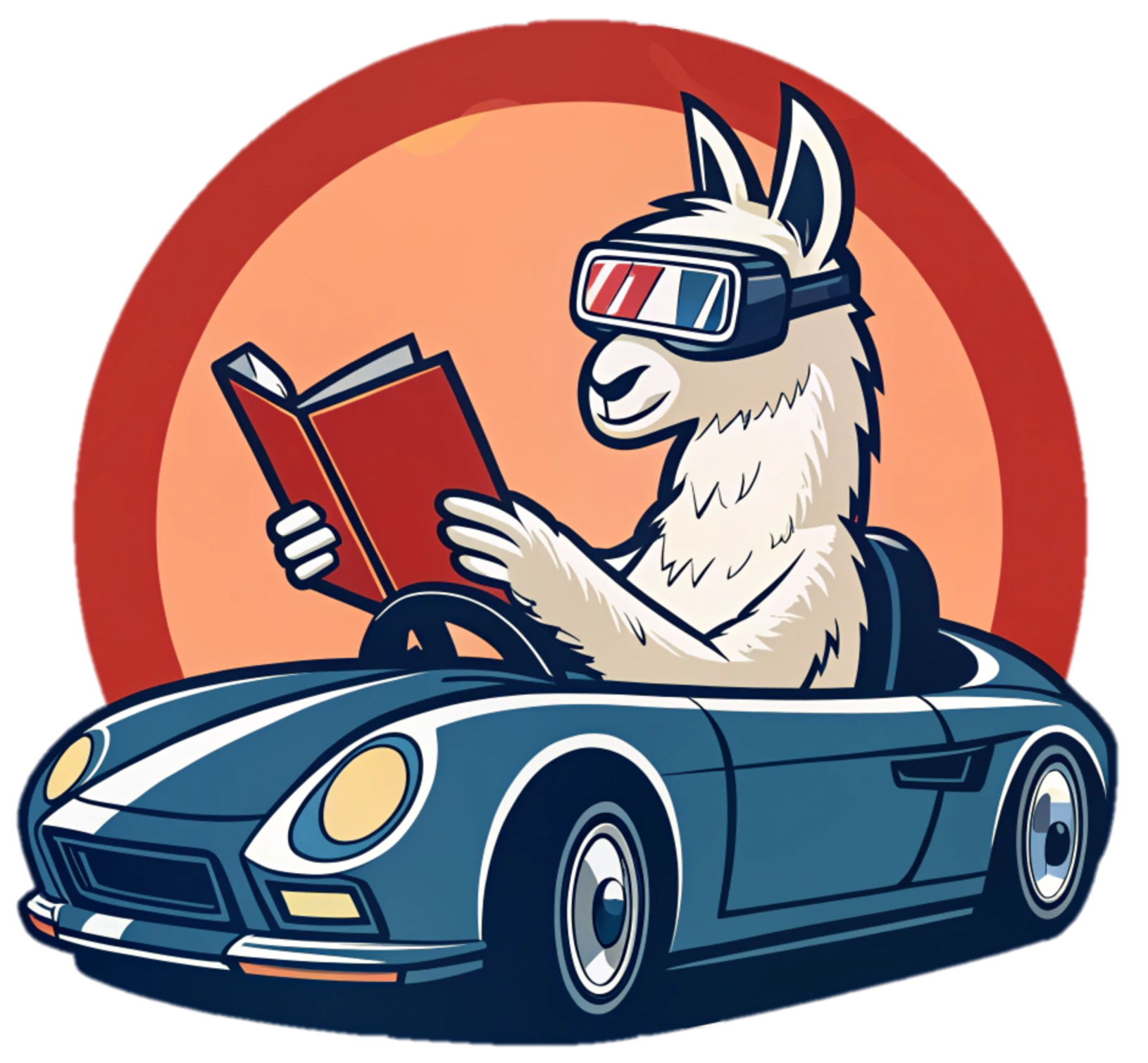}}
JarvisIR: Elevating Autonomous Driving Perception \\with Intelligent Image Restoration}

\author{
Yunlong Lin$^{1 \scalebox{0.75}{*} \scalebox{0.6}{$\clubsuit$}}$ \quad Zixu Lin$^{1 \scalebox{0.75}{*}\scalebox{0.6}{$\clubsuit$}}$ \quad Haoyu Chen$^{2\scalebox{0.75}{*}}$ \quad Panwang Pan$^{3 \scalebox{0.75}{*}}$ \quad Chenxin Li$^{6}$ \\ \quad Sixiang Chen$^{2}$ \quad
Yeying Jin$^{4}$ \quad Wenbo Li$^{5 \dag}$ \quad Xinghao Ding$^{1 \dag}$ \\
$^{1}$ Key Laboratory of Multimedia Trusted Perception and Efficient Computing, \\Ministry of Education of China, Xiamen University, Xiamen, Fujian, China \quad \\
$^{2}$ The Hong Kong University of Science and Technology (Guangzhou) \\
$^{3}$ Bytedance’s Pico \quad
$^{4}$ Tencent \quad
$^{5}$ Huawei Noah's Ark Lab \\
$^{6}$ The Chinese University of Hong Kong \\
{\tt\small Project page: \url{https://cvpr2025-jarvisir.github.io/}}
\vspace{-0.7cm} 
}

\definecolor{customblue}{HTML}{E7EFFA}
\definecolor{custompink}{HTML}{F7E1ED}
\begin{document}
\maketitle
\footnote{
$*$ Authors Yunlong Lin and Zixu Lin contributed the most and led the study, while Authors Haoyu Chen and Panwang Pan also made significant contributions. 
\quad $\dagger$ Corresponding author.}

\begin{abstract}
Vision-centric perception systems struggle with unpredictable and coupled weather degradations in the wild. Current solutions are often limited, as they either depend on specific degradation priors or suffer from significant domain gaps. To enable robust and autonomous operation in real-world conditions, we propose JarvisIR, a VLM-powered agent that leverages the VLM as a controller to manage multiple expert restoration models. To further enhance system robustness, reduce hallucinations, and improve generalizability in real-world adverse weather, JarvisIR employs a novel two-stage framework consisting of supervised fine-tuning and human feedback alignment. Specifically, to address the lack of paired data in real-world scenarios, the human feedback alignment enables the VLM to be fine-tuned effectively on large-scale real-world data in an unsupervised manner. To support the training and evaluation of JarvisIR, we introduce CleanBench, a comprehensive dataset consisting of high-quality and large-scale instruction-responses pairs, including \textbf{150K} synthetic entries and \textbf{80K} real entries. Extensive experiments demonstrate that JarvisIR exhibits superior decision-making and restoration capabilities. Compared with existing methods, it achieves a \textbf{50\%} improvement in the average of all perception metrics on CleanBench-Real. Project page: \url{https://cvpr2025-jarvisir.github.io/}.

\end{abstract}
\vspace{-0.6cm}
\section{Introduction}
\begin{figure*}[!t]
    \centering
\setlength{\abovecaptionskip}{0.1cm} %调整caption与图的距离
    \includegraphics[width=0.96\linewidth]{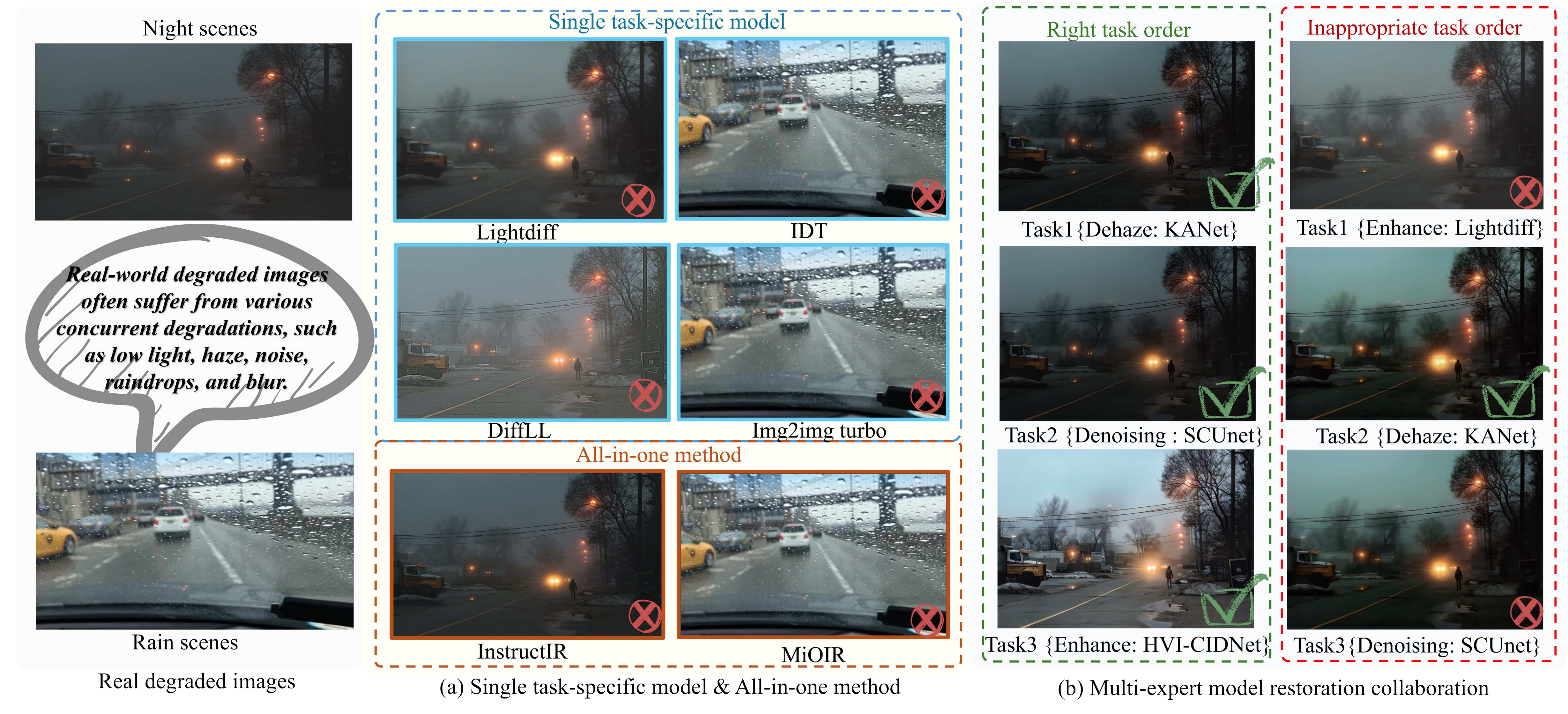}
    \caption{\textbf{Limitations of single-task methods, all-in-one methods, and inaccurate task order}. (a) Single-task specific and all-in-one methods fail to address coupled degradation in real-world scenarios. (b) Collaboration among multi-expert models effectively mitigates complex degradation, but is sensitive to the order of tasks. Unlike these approaches, JarvisIR can dynamically schedule different expert models in response to the rapidly changing scenarios and coupled degradation in the wild.}
    \label{intro}
\end{figure*}

Vision-centric perception systems often struggle in adverse weather, where images captured in real-world scenarios exhibit multiple and coupled degradations. 
Current adverse weather image restoration methods are primarily categorized into task-specific methods and all-in-one approaches. Both categories struggle with real-world coupled degradations, as shown in Figure~\ref{intro}. Task-specific methods~\cite{ye2022perceiving,jin2022structure,he2025reti,jin2022unsupervised,lin2024aglldiff} often require prior knowledge of specific degradation types, while real-world degradations are often unknown and coupled. All-in-one methods~\cite{jiang2023autodir,luo2023controlling,kong2024towards,conde2024high,chen2025teaching} trained on synthetic datasets in a supervised manner, suffer from a significant domain gap when applied to real-world data. One promising strategy to tackle multiple degradations in the wild is to integrate specialized models that excel in their domains. However, this strategy is highly sensitive to task order, and even minor changes in execution sequence can lead to significant performance degradation. Therefore, autonomously and efficiently coordinating expert models in real-world scenarios is essential for perceptual restoration.

Recently, large language models (LLMs) have exhibited remarkable proficiency in reasoning, decision-making and interaction with environments~\cite{patil2023gorilla,jain2022hugging,yang2024gpt4tools,zhao2025see,jiang2024med}. These advancements raise an important question: \textit{Could vision-language models (VLMs) act as controllers, managing publicly available specialized restoration models, autonomously planning tasks, and selecting models to facilitate the development of comprehensive restoration systems?} The answer is affirmative, however, constructing such systems is non-trivial and typically requires extensive paired data. In real-world scenarios, while there exists extensive real degraded data, the lack of corresponding labels prevents the implementation of supervised fine-tuning approaches. To tackle this issue and harness large-scale unlabeled data, we design a fine-tuning framework based on human feedback, allowing the VLM to be trained in an unsupervised manner. With this approach, we could create a system that performs robustly and reliably in the wild.

In this work, we introduce JarvisIR, a VLM-powered agent integrating VLM (i.e., Llava-Llama3~\cite{liu2024visual}) with expert restoration models sourced from GitHub and Hugging Face. The development of this system involved two key components: 1) CleanBench, an instruction-following dataset constructed using the self-instruct strategy~\cite{wang2022self}, which includes 150K synthetic and 80K real instruction-response pairs (CleanBench-Real), designed to support both training and evaluation.
2) A supervised fine-tuning (SFT) and human feedback alignment framework for training a VLM as an agent to be reliable and autonomous. Specifically, to enable the VLM to follow user instructions and perceive image degradation, we train it using the synthetic portion of CleanBench via SFT~\cite{ouyang2022training}. To enhance system robustness, reduce hallucinations, and improve generalizability in real-world adverse weather, we fine-tune JarvisIR on CleanBench-Real with human feedback. To ensure stability during training and improve overall performance, we propose the MRRHF algorithm, an extension of the ranking responses with human feedback (RRHF) approach~\cite{yuan2023rrhf}. Specifically, to expand the exploration space while maintaining a performance lower bound for JarvisIR, we introduce a hybrid sample generation strategy and regularization term. Furthermore, to comprehensively feedback the quality of system responses during training, we incorporate multiple VLM-based Image Quality Assessment (IQA) models as a unified reward model.

Our contributions can be summarized as follows:
\begin{itemize}
\item We introduce JarvisIR, a VLM-powered agent that autonomously manages and coordinates multiple expert restoration models to address coupled weather degradations in real-world environments.
\item We present CleanBench, the first high-quality instruction-following dataset specifically curated for developing intelligent restoration systems, containing 150K synthetic and 80K real instruction-response pairs.
\item We propose a novel two-stage framework combining supervised fine-tuning and human feedback alignment to enhance system robustness, reduce hallucinations and improve generalizability in the wild.
\item Our experiments demonstrate that JarvisIR outperforms strong baselines in terms of decision-making and perception restoration.
\end{itemize}

\section{Related Work}
\textbf{Tool-Augmented LLMs.}
Recent studies~\cite{shen2024hugginggpt,song2024moviechat,patil2023gorilla,qin2023toolllm,schick2024toolformer,zhao2024diffagent,chai2024auroracap} highlight the growing potential of large language models (LLMs) for proficient tool usage and decision-making in complex settings. For example, Gorilla~\cite{patil2023gorilla} facilitates LLMs' response to Tool calls through dataset construction and fine-tuning. ToolLLM~\cite{qin2023toolllm} extends this concept to enable interaction with a large number of tools. ToolAlpaca~\cite{tang2023toolalpaca} demonstrates the feasibility of generalized tool-use capabilities in smaller LLMs. Toolformer~\cite{schick2024toolformer} constructs tool-use augmented data to train LLMs to select tools. In the realm of visual tools, various approaches have been proposed to enhance the capabilities of large language models in handling visual tasks~\cite{wu2023visual,yang2023mm}, augmented with Hugging Face models~\cite{shen2024hugginggpt}, Azure models~\cite{yang2023mm}, visual foundation models~\cite{wu2023visual}. 
% These works have primarily focused on a general range of API calls that only allow for limited and fixed options for a single task.

\noindent\textbf{Alignment of LLMs.}
Reinforcement Learning from Human Feedback (RLHF)~\cite{achiam2023gpt,anthropic2024claude,team2023gemini,jiang2024modality} has emerged as a groundbreaking technique for aligning LLMs. The core idea is learning a reward function to reflect human preferences with human annotations and optimize LLMs by RL methods like proximal policy optimization (PPO).
During PPO-based optimization, updating LLMs requires the likelihood of an entire generation. However, for LLM agents, human feedback is usually obtained only after the tool response is completed and the function is successfully invoked. Moreover, unlike typical LLM training, our two-stage fine-tuning process integrates both visual and linguistic modalities.
Rank Responses to Align Human Feedback (RRHF)~\cite{yuan2023rrhf} has shown promise by using reward models to rank multiple responses, aligning LLMs effectively. This technique allows easy extension to fine-grained tool agents, thereby maximizing the utility of existing reward models.

\noindent\textbf{Image Restoration.}
Single-task image restoration has achieved significant progress in addressing specific degradation types, such as dehazing \cite{wu2023ridcp,jin2022structure,lin2024nighthaze}, low-light enhancement \cite{jiang2024lightendiffusion,jin2023enhancing,lin2024unsupervised}, desnowing \cite{chen2021all,cheng2023snow,Chen_2023_ICCV}, deraining \cite{xiao2022image,chen2023learning,chen2023sparse}, denoising~\cite{zhang2023practical,chen2023masked} super-resolution~\cite{wang2021real,chen2024low,wang2024exploiting,sun2024coser},  image fusion~\cite{he2023degradation,wang2025learning,he2023hqg,lin2023domain}. However, these task-specific approaches often lack generalizability and adaptability to complex, coupled degradations. To overcome this limitation, adverse weather restoration research aims to develop a unified framework capable of addressing multiple degradation types simultaneously \cite{chen2022learning,li2020all,ozdenizci2023restoring,he2024diffusion}.  Another prevailing research line is dedicated to building more intelligent restoration systems. Clarity ChatGPT~\cite{wei2023clarity}, integrated with advanced visual models, allows users to perform sophisticated image manipulation and enhancement through natural language interactions. RestoreAgent~\cite{chen2024restoreagent} and AgenticIR~\cite{zhu2024intelligent} \yeying{are contemporaneous independent works that utilize MLLM as a task planner to coordinate multiple restoration tools.} Specifically, RestoreAgent~\cite{chen2024restoreagent} involves fine-tuning a vision-language model (VLM) using synthetic datasets to directly produce an execution plan. AgenticIR~\cite{zhu2024intelligent} leverages two off-the-shelf LLMs and VLMs to achieve the scheduling of restoration tools on the synthetic experiment platform. Essentially, both studies focus on building intelligent restoration systems tailored for synthetic degradation conditions.
\textit{Conversely, our study aims to develop a robust system for real-world applications, incorporating human feedback to enhance robustness, reduce hallucinations and improve generalizability. Furthermore, \yeying{our} system is built in an unsupervised manner using large-scale, unlabeled real-world data.}

\begin{figure}[!t]
    \centering
\setlength{\abovecaptionskip}{0.1cm} %调整caption与图的距离
    \setlength{\belowcaptionskip}{-0.3cm}%调整caption与下文的距离
    \includegraphics[width=1\linewidth]{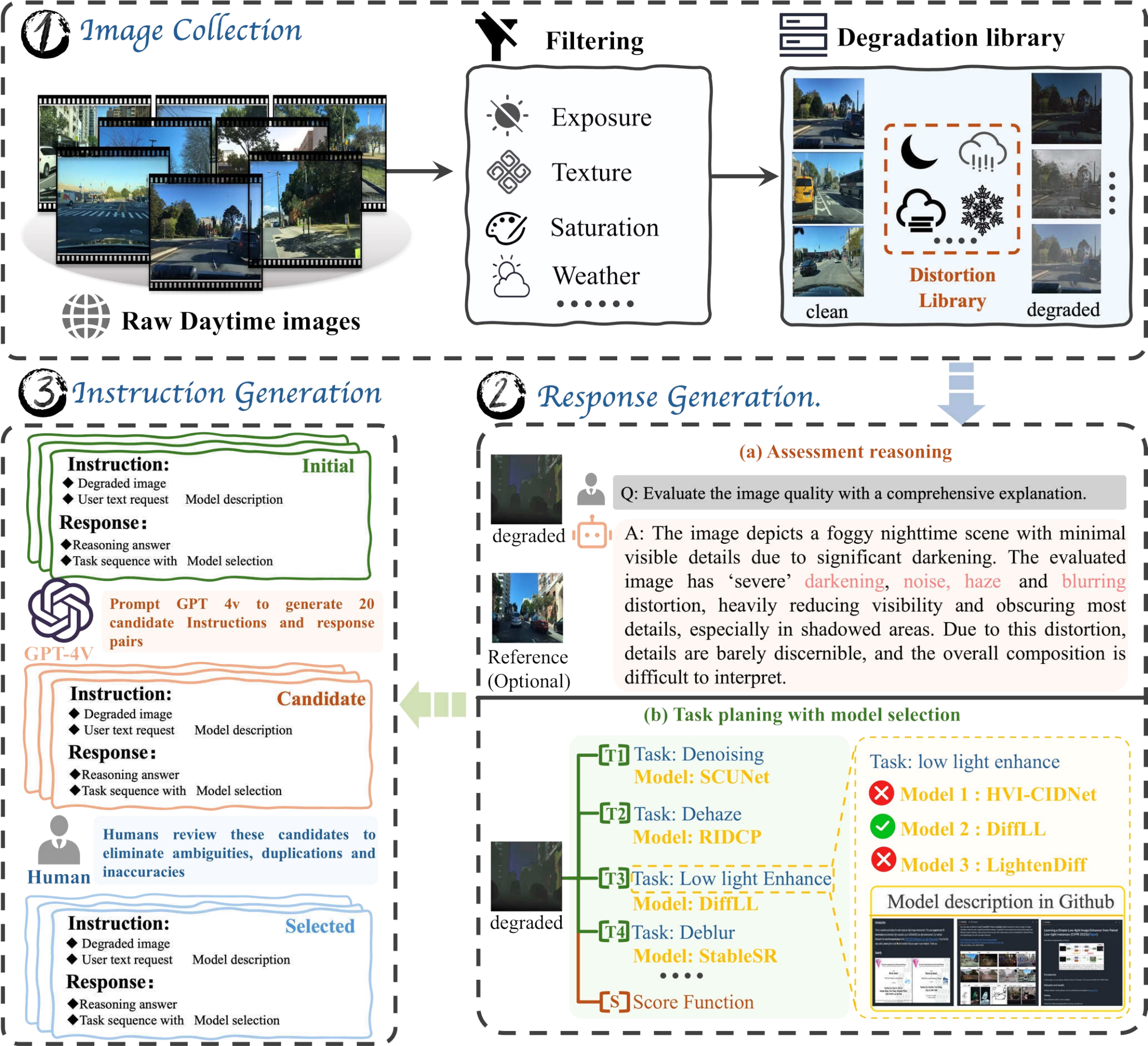}
    \caption{The dataset construction workflow consists of three main steps: 1) Synthesis of degraded images. 2) Generation of Assessment reasoning and the optimal task sequence. 3) Generation of instruction-response pairs for the system.
    }
    \label{data_generated}
\end{figure}
% \section{Task Formulation}
% instruct-
\section{Methodology}

In this section, we first describe CleanBench, a comprehensive benchmark consisting of extensive instruction-response pairs used for the training and evaluation of JarvisIR (Sec.~\ref{subsec:CleanBench}).
We then introduce JarvisIR, a VLM agent to call expert restoration models in response to intricate multiple degraded environments in the wild (Sec.~\ref{subsec:JarvisIR}). Finally, we describe the two-stage training framework for JarvisIR, comprising supervised fine-tuning and human feedback alignment.

\subsection{CleanBench}
\label{subsec:CleanBench}
High-quality and large-scale datasets are crucial for unleashing the full potential of VLMs. A multimodal instruction sample can be formally represented as a triplet: \{\textit{user instruction, degraded image, response}\}, where ``\textit{user instruction}" specifies the task and describes the restoration tools, ``\textit{degraded image}" serves as the visual input to be processed, and the ``\textit{response}" provides the ground truth answer. In Figure~\ref{data_generated}, we outline the construction of our dataset, focusing on the generation of degraded images and the collection of task-specific instructions and responses.

\noindent\textbf{Image Collection.} We first collect raw daytime images from various sources, including autonomous driving datasets~\cite{caesar2020nuscenes,yu2020bdd100k,sakaridis2021acdc,zurn2024wayvescenes101} and natural scenes~\cite{yang2021sparse,cai2018learning,zhou2022lednet,jin2025raindrop,lin2024nighthaze,li2018benchmarking,liu2018desnownet}. Then, Q-instruct~\cite{wu2024q} serves as a quality filter to extract high-quality samples. To simulate realistic adverse weather scenarios, including rainy, nighttime, snowy, and foggy, we customized the degradation library developed using physical models and image transformation techniques to synthesize degraded images. More detail in supplementary material.

\begin{figure}[!t]
    \centering
\setlength{\abovecaptionskip}{0.1cm} %调整caption与图的距离
    \setlength{\belowcaptionskip}{-0.5cm}%调整caption与下文的距离
    \includegraphics[width=0.96\linewidth]{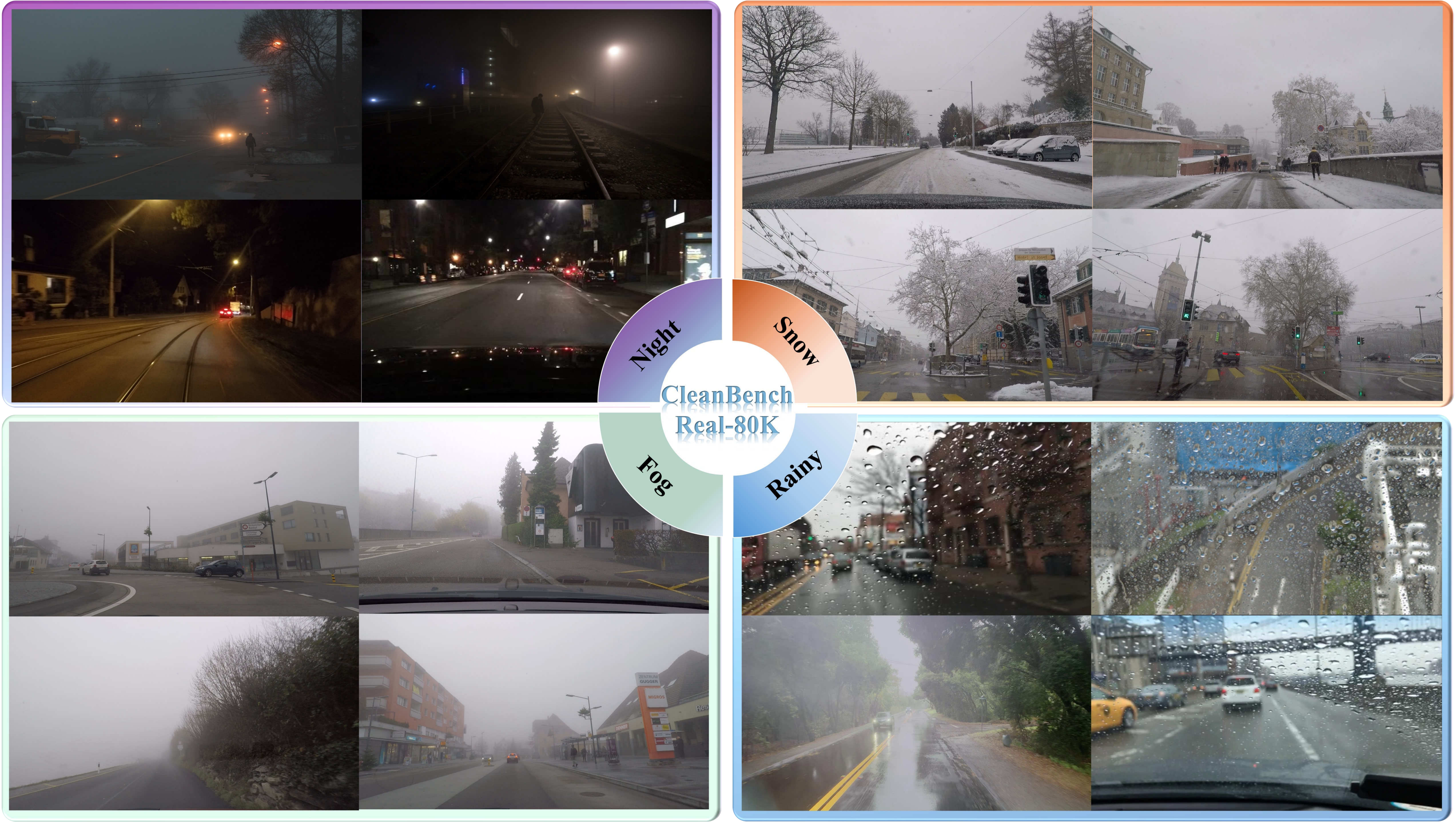}
    \caption{Examples of CleanBench-Real dataset.}
    \label{cleanBench_example}
\end{figure}

\noindent\textbf{Response Generation.} The response from JarvisIR consists of two components: ``chain-of-thought" (COT) rationales and the optimal task sequence with model selection. (a) For COT rationales, we distill DepictQA-Wild's~\cite{you2024descriptive} knowledge, which excels in low-level quality reasoning assessment. Specifically, given a degraded image pair, we prompt DepictQA-Wild~\cite{you2024descriptive} to assess the quality of the degraded image in terms of clarity, colorfulness, and sharpness, generating detailed degradation and reasoning insights. (b) To determine the optimal task sequence with restoration model selection, we employ an exhaustive search strategy~\cite{chen2024restoreagent} to explore various task permutations and model combinations, scoring each sequence to identify the optimal restoration path. 

\noindent\textbf{Task-model Assignment.} 
User instructions include descriptions of available tasks and models, sourced from GitHub or Hugging Face, to formulate task-model assignment as a single-choice problem. Presenting tasks and models as options within a context allows JarvisIR to more effectively identify the appropriate model for each sub-task.

\noindent\textbf{Instruction Generation.}  
Motivated by the self-instruct strategy~\cite{wang2022self}, for each initial user instruction and response, GPT-4V is prompted to generate 20 candidate pairs. We then manually review these candidates to eliminate ambiguity, repetition, and inaccuracies, ultimately selecting 5 instruction-response pairs per degraded image (see supplementary material for details). Ultimately, CleanBench includes a total of \textit{150K} instruction-response pairs, which are used in the initial instruction-tuning phase.

\noindent\textbf{CleanBench-Real.}
To align and evaluate JarvisIR's performance in real-world scenarios, we introduce CleanBench-Real, comprising \textit{80K} unlabeled real degraded images from internet and diverse sources~\cite{caesar2020nuscenes,yu2020bdd100k,wang2013naturalness,jin2025raindrop,jin2023enhancing,yang2021sparse,liu2018desnownet,quan2021removing}. CleanBench-Real is categorized into four adverse weather scenarios: rainy, night, snowy, and foggy. The degradation in each scenario is complex and interwined. For example, as presented in Figure~\ref{cleanBench_example}, an image captured in rain may experience multiple degradations concurrently, including rain, raindrops, defocus blur, and noise (more in supplementary material). For the division of the training and evaluation sets, we selected 500 images from each of the four CleanBench-Real scenarios to form the evaluation set (\textit{2K}), while the remaining images are utilized for alignment tuning. Instruction-response pairs are generated in the same way as outlined in CleanBench.

\begin{figure}[!t]
    \centering
\setlength{\abovecaptionskip}{0.1cm} %调整caption与图的距离
    \setlength{\belowcaptionskip}{-0.5cm}%调整caption与下文的距离
    \includegraphics[width=1\linewidth]{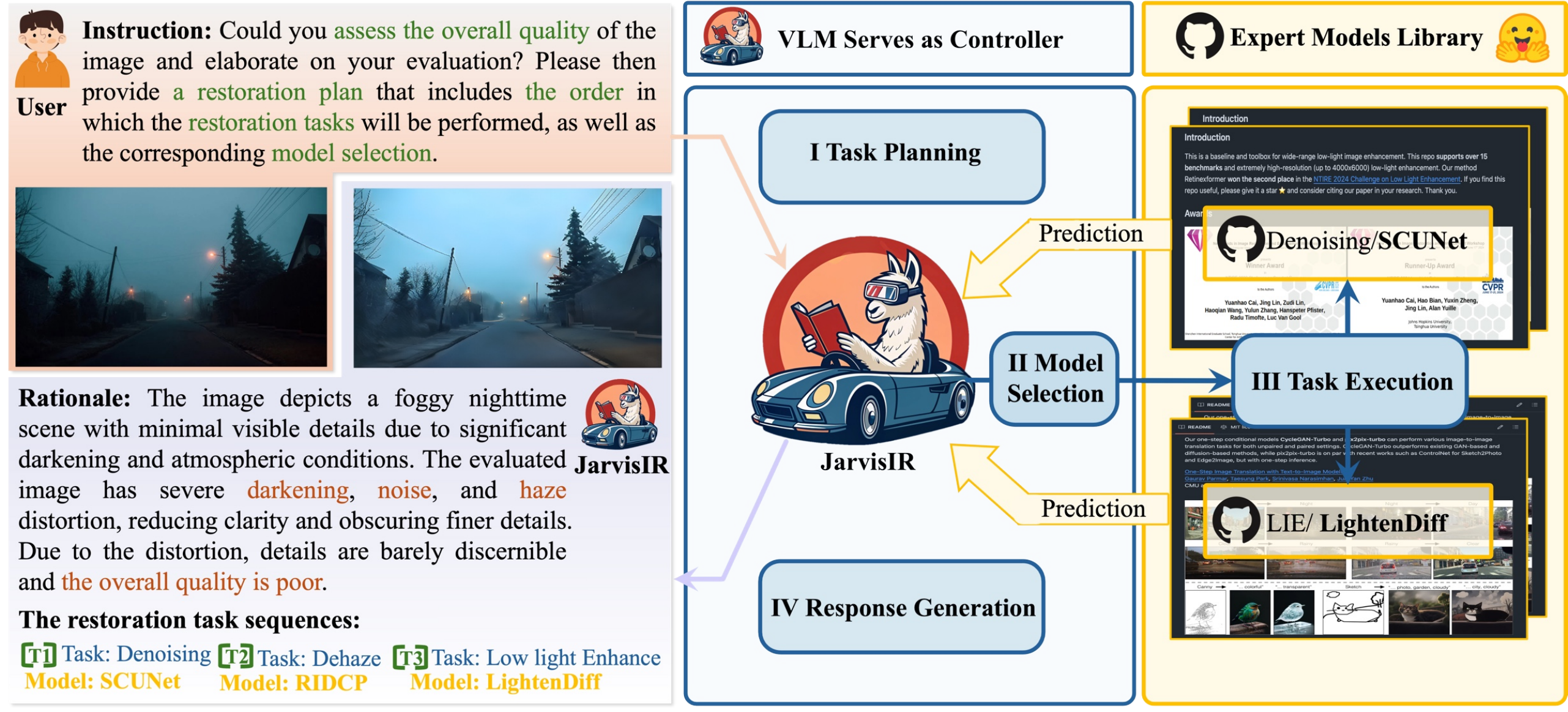}
    \caption{The workflow of JarvisIR. To address real-world coupled weather degradation, we develop JarvisIR, a VLM-powered intelligent system that dynamically schedules expert models for restoration. Initially, JarvisIR assesses the degradation of the input images and parses user instructions to formulate a task plan, selecting the appropriate expert models for each subtask. The selected experts perform their designated tasks and return the results to JarvisIR, which integrates the outcomes and provides the final answer to the user. The design of the figure is inspired by~\cite{shen2024hugginggpt}.}
    \label{workflow}
\end{figure}

\subsection{JarvisIR}
\label{subsec:JarvisIR}
\begin{figure}[!t]
    \centering
\setlength{\abovecaptionskip}{0.1cm} %调整caption与图的距离
    \includegraphics[width=1\linewidth]{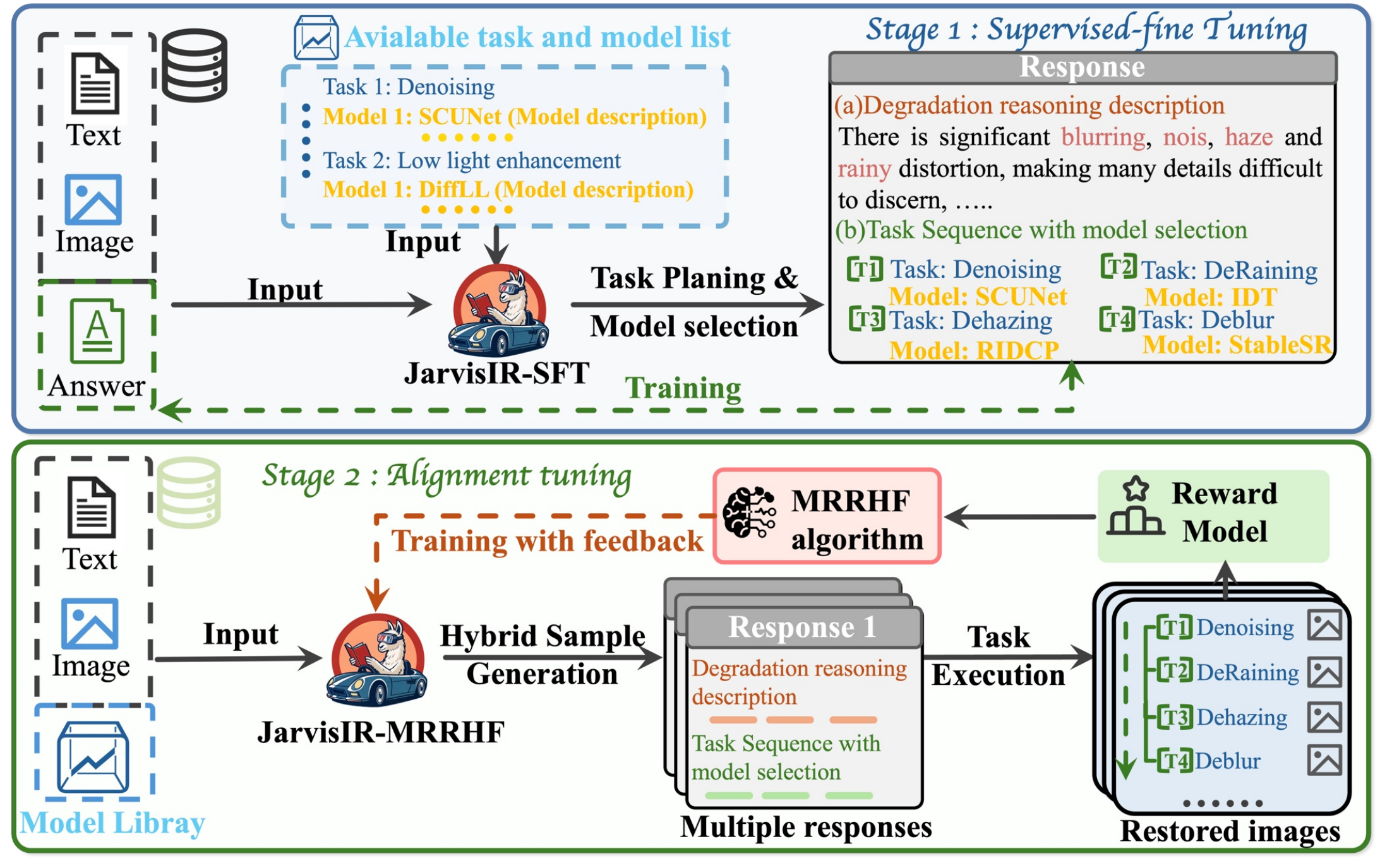}
    \caption{Two-stage training framework of JarvisIR. In the first stage, JarvisIR undergoes supervised fine-tuning on synthetic data from CleanBench to enable it to follow user instructions and recognize image degradation. In the second stage, we further fine-tune JarvisIR on CleanBench-Real using the MRRHF algorithm to improve system robustness, reduce hallucinations, and enhance generalizability under real-world adverse weather conditions.}
    \label{framework}
\end{figure}

JarvisIR is a VLM-powered agent that coordinates multiple expert restoration models to address complex degradation. As illustrated in Figure~\ref{workflow}, the workflow of JarvisIR consists of four steps: Task Planning, Model Selection, Task Execution, and Response Generation. To enhance the agent's decision-making and perception restoration capabilities in real-world scenarios, as depicted in Figure~\ref{framework}, we initially perform supervised fine-tuning (SFT) on CleanBench to obtain an initial version, termed JarvisIR-SFT. Subsequently, the JarvisIR-SFT is further fine-tuned utilizing the MRRHF algorithm on CleanBench-Real, yielding the JarvisIR-MRRHF model.

\subsubsection{JarvisIR-SFT}
\label{subsec:SFT}
We employ the standard SFT to get the JarvisIR-SFT model. Formally, the multimodal instruction sample can be denoted in a triplet form $(\mathcal{I}, \mathcal{M}, \mathcal{R})$, where $\mathcal{I}$, $\mathcal{M}$, $\mathcal{R}$ represent the user instruction, the degraded image, and the ground truth response, respectively. The VLM predicts an answer $\mathcal{A}$ given the instruction and the degraded image: $\mathcal{A}=f(\mathcal{I}, \mathcal{M} ; \theta)$. The training objective is the original auto-regressive objective used to train LLMs~\cite{luo2024cheap,yin2023survey}:
\begin{equation}
L_{sft}=-\sum_{i=1}^N{\log}P_{\pi}\left( \mathcal{R} _i\mid \left\{ \mathcal{I}_i ,\mathcal{M}_i \right\} ,\mathcal{R} _{<i};\theta \right) ,
\end{equation}
where $N$ is the length of the ground-truth response.

\subsubsection{JarvisIR-MRRHF}
\label{subsec:MRRHF}
Intuitively, SFT allows JarvisIR-SFT to achieve favorable performance on synthetic data. Nevertheless, as previously noted, due to the distribution shift, transferring from synthetic training data to real test data, JarvisIR-SFT exhibits increased hallucination, i.e., degraded perception restoration performance and decision-making capability. To improve its generalizability, we further fine-tune JarvisIR on CleanBench-Real with refined ranking responses with human feedback algorithm (MRRHF).

\noindent\textbf{Reward modeling.} The reward model evaluates tool-calling outcomes and converts them into structured reward signals to guide the agent's optimization process. Therefore, selecting an appropriate reward model is crucial. Fortunately, in the image quality assessment (IQA) field, VLM-based IQA models have been developed~\cite{wu2024q}, demonstrating strong performance in evaluating aesthetic quality and image distortion. These IQA models are inherently suitable for serving as reward models. To construct a comprehensive reward model $\mathcal{S}$, as well as an evaluation system, we integrated multiple IQA models. Specifically, we employ a z-score strategy~\cite{chen2024restoreagent} to standardize the scores assessed by each IQA model separately and then sum the standardized results:
\begin{equation}
\mathcal{S}=\sum_{i=1}^k{\frac{s_i-\mu _i}{\sigma _i}},
\label{reward_model}
\end{equation}
where $s_i$ represents the score assessed by $i$-th IQA model. $\mu_i$ and $\sigma_i$ epresent the mean and standard deviation of $s_i$, respectively. $k$ indicates the total number of IQA models.

\noindent\textbf{Alignment with MRRHF.} We propose an extension to the existing RRHF method that can be used for aligning JarvisIR in a cost-effective manner: 1) A hybrid sample generation strategy that combines offline and online approaches to expand the optimization exploration space while ensuring a performance lower bound. 2) Entropy regularization terms are integrated to foster diversity among agent responses, thereby facilitating exploration during training. Specifically, for a pair of user instruction $\mathcal{I}_i$ and degraded image $\mathcal{M}_i$, we first adopt offline diverse beam search~\cite{vijayakumar2016diverse} to get $m_1$ different responses $\mathcal{R}_{m_1}=\left\{r_1, r_2, \ldots, r_{m_1}\right\}$ from SFT model $\pi$. Similarly, we can obtain $\mathcal{R}_{m_2}=\left\{r_1, r_2, \ldots, r_{m_2}\right\}$ from policy model $\rho$ (initialized from SFT model $\pi$) during training. The combined candidate $m$ responses are denoted as $\mathcal{R}_{m}=\mathcal{R}_{m_1}\cup \mathcal{R}_{m_2}$. Subsequently, we execute the task sequences specified in candidate responses, calling multiple restoration models to generate restored images. These predictions are then assessed by the reward model $\mathcal{S}$, yielding scores for each $r_i$ with $\mathcal{S}(r_i)=s_i$.
To align with scores $\left\{ s_i \right\} _m$, we use policy model $\rho$ to give scores $p_i$ for each $r_i$ by:
\begin{equation}
p_i=\frac{\sum_t{\log}P_{\rho}\left( r_{i,t}\mid \left\{ \mathcal{I} _i,\mathcal{M} _i \right\} ,r_{i,<t};\theta \right)}{\left\| r_i \right\|},
\end{equation}
where $p_i$ is conditional log probability (length-normalized) of $r_i$ under model $\rho$. The core idea is letting the policy model $\rho$ give larger probabilities for better responses and give smaller probabilities for worse responses. Inspired by PRO~\cite{song2024preference}, we refine the original ranking loss:
\begin{equation}
L_{\mathrm{rank}}=\sum_{s_i<s_j}{\left( s_j-s_i \right) \max}\left( 0,p_i-p_j \right) ,
\end{equation}
and a cross-entropy loss like SFT process is added to learn the response with the highest reward $s_i$, $i^{\prime} = \mathrm{arg}\max_i s_i$:
\begin{equation}
% \begin{split}
L_{ft} = -\sum_t \log P_{\rho}\left( r_{i^{\prime},t} \mid \left\{ \mathcal{I}_i, \mathcal{M}_i \right\}, r_{i^{\prime},<t}; \theta \right).
% \end{split}
\end{equation}
Furthermore, we define the entropy regularization term as:
\begin{equation}
    L_{er} = -\sum_a \rho(a \mid y) \log \rho( a\mid y),
\end{equation}
where $y$ represents the current state of the agent. The overall loss is utilized to optimize the JarvisIR-SFT to derive JarvisIR-MRRHF:
\begin{equation}
    L=\lambda _1 L_{r a n k}+\lambda _2 L_{ft}+\lambda _3 L_{er},
\end{equation}
where $\lambda _1$, $\lambda _2$ and $\lambda _3$ are constants controlling the relative importance of the different losses, which are empirically set to 0.5, 0.5 and 0.1 in all experiments, respectively.

\begin{figure*}[!t]
    \centering
\setlength{\abovecaptionskip}{0.1cm} %调整caption与图的距离
    \setlength{\belowcaptionskip}{-0.4cm}%调整caption与下文的距离
    \includegraphics[width=1\linewidth]{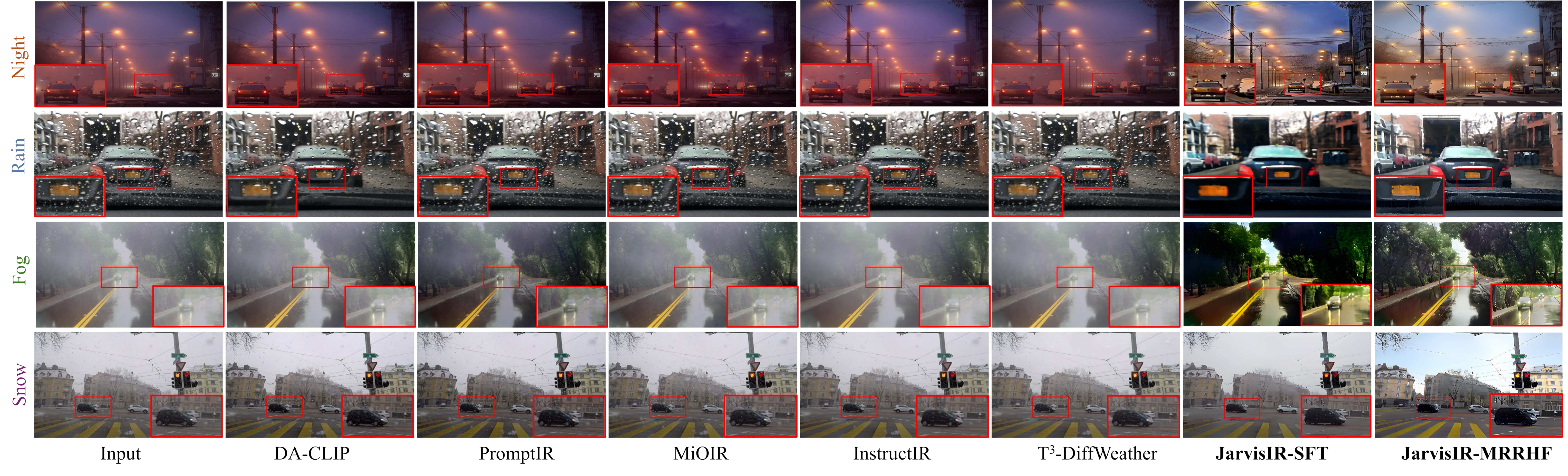}
    \caption{Visual comparisons of various methods on CleanBench-Real. Our approach delivers significant quality improvements, eliminating complex real-world degradation and preserving the most natural details.}\label{perception_results}
\end{figure*}
% call APIs to retrieve the images and use Eq. (5) as the scoring function for the reward model to get n scores for each ri with
\noindent\textbf{\textcolor[RGB]{116, 174, 212}{Discussion of RLHF and RRHF:}} The training of vanilla RLHF~\cite{ouyang2022training} necessitated a policy model, a value model, a reward model, and a reference model, which could be demanding on memory resources. Rank Responses to Align Human Feedback (RRHF)~\cite{yuan2023rrhf} can effectively alleviate the issues of resource-intensive and tedious hyperparameter tuning in RLHF. However, directly fine-tuning JarvisIR using RRHF yields limited improvement to its generalization in real-world scenarios. Although vanilla RRHF employs an off-policy learning strategy that could save time by avoiding the need to generate new responses during training, it has the drawback of relying on a static offline preference dataset for training the policy model. Consequently, the policy might over-optimize for reward on in-distribution data as the model cannot further query the preference oracle during the training process~\cite{wang2024comprehensive}. The RRHF incorporating online sampling like PPO might mitigate this issue, but it demands more GPU resources to store the reference model, thereby significantly decreasing the training speed~\cite{yuan2023rrhf}.

\section{Experiments}
\subsection{Experimental Settings}
\textbf{Training Setup.} Llava-Llama3-8b~\cite{liu2024visual} serves as the base model for JarvisIR, which undergoes full parameter fine-tuning using the Adam optimizer. During the SFT phase, we fine-tune JarvisIR for 3 epochs with a batch size of 128 and a learning rate of 1e-5. In the MRRHF tuning phase, we set the diverse beam search size to 3, the diverse beam group to 5, the diversity penalty to 2.0, and the sampling temperature to 0.8. Alignment tuning is performed over 3 epochs with a batch size of 1 and a learning rate of 1e-5. To speed up training, we select three IQA models—Q-instruct~\cite{wu2024q}, MUSIQ~\cite{ke2021musiq} and MANIQA~\cite{yang2022maniqa}—to construct the unifed reward model (Eq.~\ref{reward_model}). All experiments are conducted on 8 NVIDIA A100 80G GPUs.

\noindent\textbf{Dataset Setings \& Metrics.} 
The CleanBench is fully utilized for supervised fine-tuning of Llava-Llama3-8b~\cite{liu2024visual} to obtain JarvisIR-SFT. The training set of CleanBench-Real is used for alignment tuning, yielding JarvisIR-MRRHF. Additionally, JarvisIR's evaluation is conducted on the validation set of CleanBench-Real, focusing on 1) decision-making ability and 2) perception restoration capability in real-world scenarios. Due to the lack of paired clean-degraded data in the real scenarios. Four image quality assessment metrics are used for evaluation: MUSIQ~\cite{ke2021musiq}, MANIQA~\cite{yang2022maniqa}, CLIP-IQA+~\cite{wang2023exploring}, LIQE~\cite{zhang2023blind}.

\noindent\textbf{Tool Settings.} We present the task-specific restoration tools employed in our implementation, including denoising (SCUnet~\cite{zhang2023practical}), super-resolution \& deblur \& compression artifact removal (StableSR-turbo~\cite{wang2024exploiting} and Real-ESRGAN~\cite{wang2021real}), deraining (IDT~\cite{xiao2022image}, UDR-S2Former~\cite{Chen_2023_ICCV} and Img2img-turbo~\cite{parmar2024one}), dehazing (RIDCP~\cite{wu2023ridcp} and KANet~\cite{feng2024advancing}), low-light enhancement (Img2img-turbo~\cite{parmar2024one}, HVI-CIDNet~\cite{yan2024you} and  LightenDiff~\cite{jiang2024lightendiffusion}) and desnowing (Img2img-turbo~\cite{parmar2024one} and Snowformer~\cite{chen2022snowformer}). More details are in the supplementary material. Notably, we select lightweight and efficient models instead of the latest state-of-the-art models to simplify the validation process of our proposed paradigm. Incorporating more advanced models could further enhance performance.

\begin{table}[t!]
\centering
\caption{Comparison of JarvisIR with other strategies on the CleanBench-Real validation set. The ``Score" represents the sum of the four normalized metrics. The ``Ranking'' indicates the given decision's percentage ranking among all possible decisions. We highlight the \colorbox{lightpeach!90}{best} and \colorbox{customblue}{second-best} results.}\label{Decision-making}
\scalebox{0.94}{
\setlength\tabcolsep{4pt}
\renewcommand\arraystretch{1}
\begin{tabular}{ccc}
\toprule
% & \multicolumn{2}{c}{noise + JPEG} & \multicolumn{4}{c}{haze + noise} & \multicolumn{4}{c}{rain + haze + noise} \\ 
% \cmidrule(lr){2-5} \cmidrule(lr){6-9} \cmidrule(lr){10-13}
Strategy & Score & Ranking(\%) \\
\midrule
(I) Random Order and Model  & 1.12 & 43.2\% \\
(II) Random Order + Predict Model & 2.66 & 34.7\% \\
(III) Random Model + Predict Order& 3.08 & 23.4\% \\
(IV) Pre-defined Order and Model& 3.94 & 22.5\% \\
(V) Human Expert & 4.85 & 18.6\% \\
% \bottomrule
\scalebox{1.5}{\color{sh_blue}{$\star$}}\textbf{JarvisIR-SFT} & \colorbox{customblue}{5.17} & \colorbox{customblue}{14.3\%} \\
\scalebox{1.5}{\color{sh_red}{$\star$}}\textbf{JarvisIR-MRRHF} & \colorbox{lightpeach!90}{6.21} & \colorbox{lightpeach!90}{4.8\%}\\
\bottomrule
\end{tabular}}
\vspace{-0.3cm}
\end{table}

\begin{table*}[t!]
\centering
\setlength{\belowcaptionskip}{-0.2cm}%调整caption与下文的距离
\caption{Comparison of JarvisIR with All-in-One methods for multi-degraded perception restoration on CleanBench-Real. We highlight the \colorbox{lightpeach!90}{best}, \colorbox{customblue}{second-best} and \colorbox{custompink}{third-best} results. Notably, all scenes represent multiple degraded weather conditions,
such as haze, low light and blur.}\label{perception_table}
\scalebox{0.84}{
\setlength\tabcolsep{8pt}
\renewcommand\arraystretch{1}
\begin{tabular}{lcccccccc}
\toprule
~ & \multicolumn{4}{c}{Night Scenes} & \multicolumn{4}{c}{Rain Scenes} \\ 
\cmidrule(lr){2-5} \cmidrule(lr){6-9} 
\multirow{-2}*{Method} & MUSIQ $\uparrow$ & MANIQA $\uparrow$ & CLIP-IQA+ $\uparrow$ & LIQE $\uparrow$  & MUSIQ $\uparrow$ & MANIQA $\uparrow$ & CLIP-IQA+ $\uparrow$ & LIQE $\uparrow$  \\
\midrule
AirNet~\cite{li2022all}      & 44.26 & 0.1889 & 0.4429 & 1.313 & 62.61 & 0.3871 & 0.5867 & 3.136  \\
AutoDIR~\cite{jiang2023autodir} & \colorbox{custompink}{47.30} & 0.1885 & 0.4341 & 1.403 & \colorbox{custompink}{63.93} & \colorbox{custompink}{0.4002} & \colorbox{custompink}{0.6082} & \colorbox{custompink}{3.312}  \\
DA-CLIP~\cite{luo2023controlling} & 45.86 & 0.2010 & 0.4544 & \colorbox{custompink}{1.427} & 63.28 & 0.3993 & 0.5959 & 3.194  \\
PromptIR~\cite{potlapalli2024promptir} & 45.45 & 0.2010 & 0.4473 & 1.408 & 62.85 & 0.3926 & 0.5941 & 3.161  \\
MiOIR~\cite{kong2024towards} & 46.93 & \colorbox{custompink}{0.2013} & 0.4403 & 1.408 & 63.07 & 0.3779 & 0.5841 & 3.055  \\
InstructIR~\cite{conde2024high} & 44.03 & 0.1533 & 0.3689 & 1.257 & 62.93 & 0.3657 & 0.5609 & 3.055  \\
T$^{3}$-DiffWeather~\cite{chen2025teaching}& 46.79 & 0.1964 & \colorbox{custompink}{0.4547} & 1.413 & 62.67 & 0.3689 & 0.5823 & 3.011 \\
\scalebox{1.5}{\color{sh_blue}{$\star$}}\textbf{JarvisIR-SFT}& \colorbox{customblue}{60.77} & \colorbox{customblue}{0.5048} & \colorbox{customblue}{0.5239} & \colorbox{customblue}{3.224} & \colorbox{customblue}{65.03} & \colorbox{customblue}{0.5339} & \colorbox{customblue}{0.6290} & \colorbox{customblue}{4.005}
  \\
\scalebox{1.5}{\color{sh_red}{$\star$}}\textbf{JarvisIR-MRRHF}& \colorbox{lightpeach!80}{67.25} & \colorbox{lightpeach!80}{0.5876} & \colorbox{lightpeach!80}{0.6336} & \colorbox{lightpeach!80}{3.613} & \colorbox{lightpeach!80}{70.38} & \colorbox{lightpeach!80}{0.7004} & \colorbox{lightpeach!80}{0.7127} & \colorbox{lightpeach!80}{4.435}
  \\
\midrule  
~& \multicolumn{4}{c}{Fog Scenes} & \multicolumn{4}{c}{Snow Scenes}\\
\cmidrule(lr){2-5} \cmidrule(lr){6-9} 
\multirow{-2}*{Method} & MUSIQ $\uparrow$ & MANIQA $\uparrow$ & CLIP-IQA+ $\uparrow$ & LIQE $\uparrow$  & MUSIQ $\uparrow$ & MANIQA $\uparrow$ & CLIP-IQA+ $\uparrow$ & LIQE $\uparrow$    \\
\midrule
AirNet~\cite{li2022all}      & 64.23 & 0.3829 & 0.6173 & 2.686 & 67.32 & \colorbox{custompink}{0.4320} & 0.6379 & 3.794  \\
AutoDIR~\cite{jiang2023autodir} & 64.84 & \colorbox{custompink}{0.3966} & 0.6443 & \colorbox{custompink}{2.928} & 67.62 & 0.4305 & \colorbox{custompink}{0.6453} & \colorbox{custompink}{3.824}  \\
DA-CLIP~\cite{luo2023controlling} & 64.78 & 0.3880 & \colorbox{custompink}{0.6540} & 2.793 & 67.71 & 0.4294 & 0.6426 & 3.817  \\
PromptIR~\cite{potlapalli2024promptir} & 64.54 & 0.3810 & 0.6417 & 2.557 & 67.34 & 0.4292 & 0.6435 & 3.776  \\
MiOIR~\cite{kong2024towards} & \colorbox{custompink}{64.93} & 0.3501 & 0.5969 & 2.415 & 67.28 & 0.4187 & 0.6404 & 3.702  \\
InstructIR~\cite{conde2024high} & 64.82 & 0.3904 & 0.6449 & 2.919 & \colorbox{custompink}{67.98} & 0.4038 & 0.6052 & 3.715  \\
T$^{3}$-DiffWeather~\cite{chen2025teaching}& 64.58 & 0.3715 & 0.6163 & 2.497 & 67.72 & 0.4129 & 0.6268 & 3.713 \\
\scalebox{1.5}{\color{sh_blue}{$\star$}}\textbf{JarvisIR-SFT}& \colorbox{customblue}{70.45} & \colorbox{customblue}{0.4855} & \colorbox{customblue}{0.6560} & \colorbox{customblue}{3.977} & \colorbox{customblue}{70.24} & \colorbox{customblue}{0.7133} & \colorbox{customblue}{0.7127} & \colorbox{customblue}{4.086}
  \\
\scalebox{1.5}{\color{sh_red}{$\star$}}\textbf{JarvisIR-MRRHF}& \colorbox{lightpeach!80}{74.22} & \colorbox{lightpeach!80}{0.7502} & \colorbox{lightpeach!80}{0.7805} & \colorbox{lightpeach!80}{4.714} & \colorbox{lightpeach!80}{73.87} & \colorbox{lightpeach!80}{0.8014} & \colorbox{lightpeach!80}{0.7918} & \colorbox{lightpeach!80}{4.881}
  \\
\bottomrule
\end{tabular}}
% \vspace{-0.1cm}
\end{table*}

\subsection{Decision Making Capability}
\textbf{Compared Baselines.} We conducted a comparative analysis of JarvisIR against several alternative approaches: (I) Random selection of both the task order and the models, assuming that task types are accurately determined. (II) Random task order, but models predicted by JarvisIR. (III) Random model selection, but task orders predicted by JarvisIR. (IV) Using a human expert's predefined order and models for different scenes, assuming the approximate scene degradation can be determined. (V) A human expert manually generates a solution case by case for each image, determining both the task sequence and the appropriate models.

\noindent\textbf{Results.} As indicated in Table~\ref{Decision-making}, strategies that involve  human expert participation—specifically settings (IV) and (V)—demonstrate strong performance compared to random strategies, ranking within the top 22.5\% and 18.6\% of all possible strategies, respectively. These results indicate the effectiveness of human experts' experience in complex decision-making processes. Interestingly, however, our JarvisIR model achieves the highest performance, surpassing even the expert-driven customization strategies. Furthermore, JarvisIR-MRRHF (4.8\%) outperforms JarvisIR-SFT (14.3\%) in both score and ranking, highlighting that the MRRHF stage in our training framework effectively mitigates hallucination errors in system responses, thereby enabling the generation of more optimal decisions.

\subsection{Perception Restoration Ability}
\textbf{Compared All-in-One Methods.} We compare JarvisIR with existing advanced all-in-one methods: AirNet~\cite{li2022all}, AutoDIR~\cite{jiang2023autodir}, DA-CLIP~\cite{luo2023controlling}, PromptIR~\cite{potlapalli2024promptir}, MiOIR~\cite{kong2024towards}, InstructIR~\cite{conde2024high}, T$^{3}$-DiffWeather~\cite{chen2025teaching}. For a fair comparison, we repeatedly run these compared methods multiple times to fully leverage their capabilities. Additionally, we supply InstructIR and AutoDIR with explicit prompts detailing degradation scenarios to optimize their performance.

\noindent\textbf{Results.} 
As shown in Table~\ref{perception_table} and Figure~\ref{perception_results}, JarvisIR outperforms existing All-in-One approaches across all metrics. In Night Scenes, JarvisIR-MRRHF achieves a MUSIQ score of 67.25, which is 42.2\% higher than AutoDIR's score of 47.30. In MANIQA, JarvisIR-MRRHF scores 0.5876, much better than DA-CLIP (0.2010) and MiOIR (0.2013). These results show that JarvisIR autonomously selects optimal task sequences and models, outperforming methods with predefined or random sequences. Additionally, JarvisIR-MRRHF also exceeds the SFT version in all scenes, with notable gains in Rain (70.38 vs. 65.03 MUSIQ) and Fog (74.22 vs. 70.45 MUSIQ). These results demonstrate that JarvisIR fine-tuned with MRRHF can improve generalizability, fewer hallucination errors, and better decision-making ability.

\section{Ablation Study}

\begin{figure}[!t]
    \centering
\setlength{\abovecaptionskip}{-0.02cm} %调整caption与图的距离
    \setlength{\belowcaptionskip}{-0.5cm}%调整caption与下文的距离
    \includegraphics[width=1\linewidth]{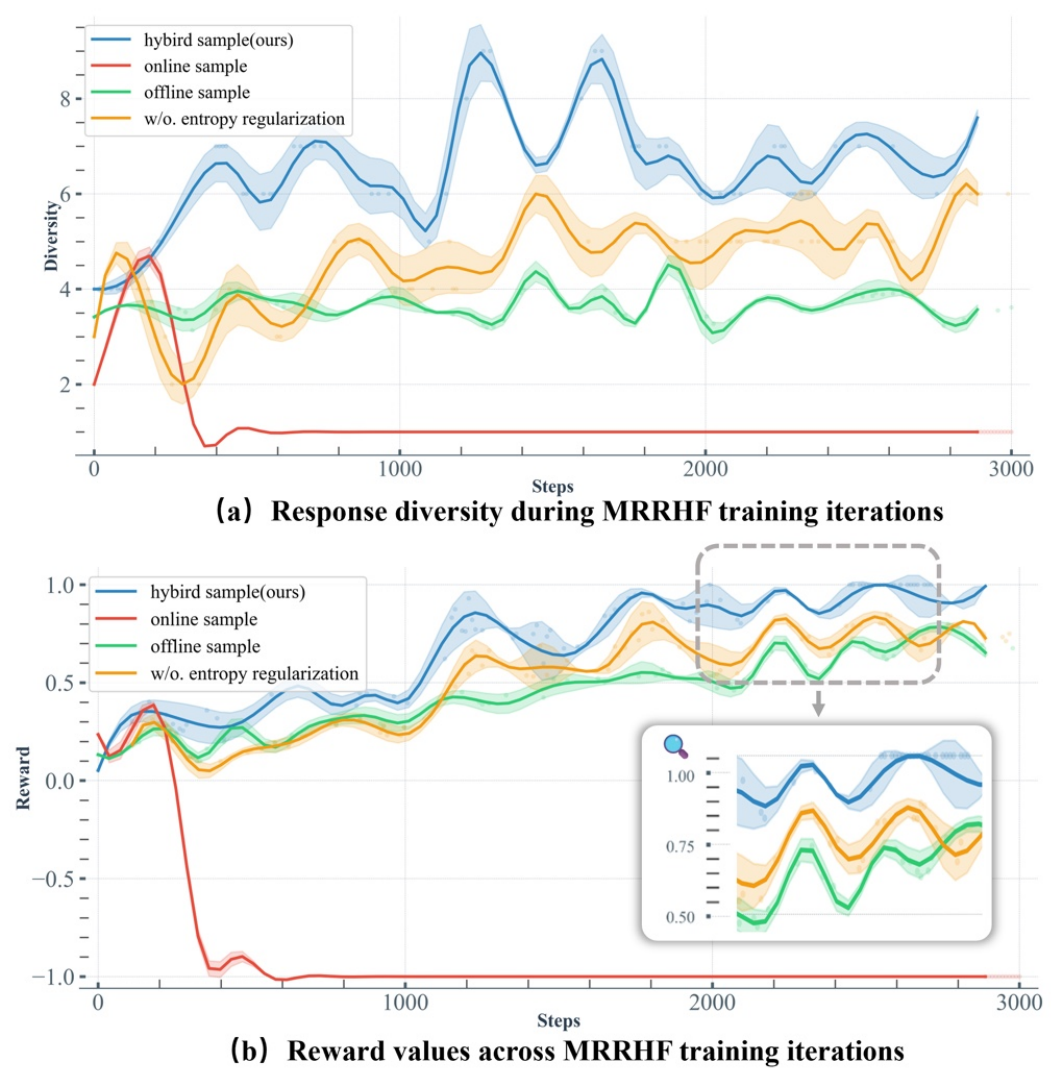}
    \caption{Ablation studies on different sample generation strategies and
entropy regularization. (a) Response diversity during MRRHF training iterations. (b) Reward values across MRRHF training iterations.}\label{ablation_results}
\end{figure}

\textbf{Sample generation strategy.} To assess the effectiveness of the hybrid sample generation strategy, we compared it with two variations of the original setting: 1) offline sample generation strategy. 2) online sample generation strategy. The results in Table~\ref{strategy} and Figure~\ref{ablation_results} yield the following observations: 1) The offline sample generation strategy yields limited performance gains, with a reward score of 0.43 and a diversity score of 3.63. This limitation arises because the sample distribution is restricted to the finite dataset generated by the SFT model using diverse beam search~\cite{vijayakumar2016diverse}. Consequently, the policy model may over-optimize for in-distribution data, thereby limiting its ability to generalize and achieve higher reward scores. 2) The online sampling strategy initially yields higher reward scores and diversity. However, as training progresses, the model encounters a collapse, leading to a significantly low reward score (-0.87) and decreased diversity (1.27). This instability may result from an excessively large optimization space without adequate constraints during training. When the model reaches a local minimum, it struggles to escape, as the candidate responses generated using diverse beam search~\cite{vijayakumar2016diverse} are of poor quality, causing the model to produce repetitive and invalid responses. Our hybrid sampling approach combines both online and offline samples, resulting in superior performance with a reward score of 0.67 and the highest diversity score of 6.55. This balanced strategy leverages the advantages of both online and offline sampling, ensuring stable training by providing sufficient exploration space while avoiding the pitfalls associated with purely online sampling. As a result, the hybrid strategy maintains high reward scores and diversity throughout training, outperforming both online and offline strategies.

% 根据这个表格的内容，帮我总结一份先进性分析的话
\begin{table}[t]
\centering
\caption{Ablation studies on different sample generation strategies and entropy regularization. The ``Reward" represents the average reward scores obtained during MRRHF training, spanning from -1 to 1. A negative score indicates a penalty, while a positive score represents a reward. The
``Diversity" reflects the average number of unique responses produced during the training process.
}\label{strategy}
\scalebox{1}{
\setlength\tabcolsep{5pt}
\renewcommand\arraystretch{1}
\begin{tabular}{ccc}
\toprule
% & \multicolumn{2}{c}{noise + JPEG} & \multicolumn{4}{c}{haze + noise} & \multicolumn{4}{c}{rain + haze + noise} \\ 
% \cmidrule(lr){2-5} \cmidrule(lr){6-9} \cmidrule(lr){10-13}
Strategy & Reward & Diversity \\
\midrule
offline sample generation & 0.43 & 3.63 \\
online sample generation & -0.87 & 1.27 \\
hybrid sample generation (ours) & \textbf{0.67} & \textbf{6.55}\\
\bottomrule
w/o. entropy regularization  & 0.50 & 4.56 \\
w. entropy regularization (ours) & \textbf{0.67} & \textbf{6.55} \\
% \textbf{JarvisIR-SFT} & 25.06 & 0.7588 \\
% \textbf{JarvisIR-RRHF} & 25.06 & 0.7588 \\
\bottomrule
\end{tabular}}
\vspace{-0.5cm}
\end{table}

\noindent\textbf{Entropy regularization.} As discussed in Sec.~\ref{subsec:MRRHF}, entropy regularization significantly affects the diversity of system responses during training. The results in Table~\ref{strategy} and Figure~\ref{ablation_results} show that without this regularization, the reward decreases from 0.67 to 0.50, while the diversity drops from 6.55 to 4.56. This highlights the role of entropy regularization in fostering greater exploration and producing more diverse, high-quality responses.

\section{Conclusions}
This paper introduces JarvisIR, a VLM-powered intelligent system that leverages Llava-Llama3 to connect distinct restoration expert models. JarvisIR can autonomously schedule different expert models in response to the rapidly changing scenarios and coupled degradation in autonomous driving and natural environments. To enhance system robustness, minimize hallucinations, and improve generalizability, we propose a novel two-stage framework comprising supervised fine-tuning and human feedback alignment. Specifically, we design the human feedback alignment to effectively tune the VLM in an unsupervised manner, leveraging large-scale unlabeled real-world data. To support the training and evaluation of JarvisIR, we present CleanBench, a high-quality, large-scale dataset containing 150K synthetic and 80K real instruction-response pairs. Experiments show that JarvisIR outperforms existing methods, achieving a 50\% improvement in the average of all perception metrics on CleanBench-Real.

\section{Acknowledgments}
This work was supported in part by the National Natural Science Foundation of China under Grant 82172033, Grant U19B2031, Grant 61971369, Grant 52105126, Grant 82272071, and Grant 62271430; and in part by the Dreams Foundation of Jianghuai Advance Technology Center; and in part by the Open Fund of the National Key Laboratory of Infrared Detection Technologies.

{
   \small
   \bibliographystyle{ieeenat_fullname}
   \bibliography{main}
}
% \input{sec/suppl_results}

% \clearpage
% \setcounter{page}{1}
\maketitlesupplementary

This is the supplementary material for the paper: ``\textit{JarvisIR: Elevating Autonomous Driving Perception
with Intelligent Image Restoration.}" We provide the following materials in this manuscript:

\begin{itemize}
    \item Sec.\ref{Tool} More implementation details.
    \begin{itemize}
        \item Restoration tool settings.
        \item Details of Model Setups.
        % \item Libraries, frameworks, and hardware specifications.
    \end{itemize}
    
    \item Sec.\ref{Datasets} CleanBench dataset details.
    \begin{itemize}
        \item Dataset statistics.
        \item Details of degradation library.
    \end{itemize}

    \item Sec.\ref{ablation} More ablation.
    \begin{itemize}
        \item MRRHF vs. vanilla RRHF.
        \item Sample generation strategy and entropy regularization.
        \item Effectiveness of differentiated contrast weights.
        \item Impact of reasoning for decision-making.
        \item Impact of reward model.
        % \item Inference time.
    \end{itemize}

    \item Sec.\ref{Results} More visual results.
    % \begin{itemize}
    %     \item Perception restoration.
        % \item Semantic segmentation and object detection.
    % \end{itemize}
    
    \item Sec.\ref{Future} Limitations, broader impacts and future work.

\end{itemize}

\section{More implementation details}\label{Tool}
\subsection{Restoration tool settings}
Table~\ref{tools_setting} lists the task-specific restoration tools used in our implementation. Notably, some models lack weights corresponding to certain tasks but are inherently adaptable; we collect appropriate data to retrain them. For example, Img2img-turbo~\cite{parmar2024one} is an image-to-image translation method based on SD-turbo that provides night-to-day and rainy-to-day weights but not snow-to-day weights. To enable Img2img-turbo to adapt to snow scenes, we retrain it using the CycleGAN paradigm on the snow subset of the ACDC dataset~\cite{sakaridis2021acdc}. Additionally, it is important to note that we are not utilizing the latest state-of-the-art tools, suggesting considerable potential for enhancing our models.

\begin{table*}[ht]
\centering
\caption{Task-specific restoration tools with descriptions.}\label{tools_setting}

\scalebox{0.96}{ % 设置缩放比例，1.0 表示正常大小
\begin{tabular}{@{}c l p{9cm}@{}}
\toprule
\textbf{Task} & \textbf{Tools} & \textbf{Model Description} \\ 
\midrule
\multirow{4}{*}{\textbf{Super-resolution}} &
\multirow{3}{*}{StableSR-turbo~\cite{wang2024exploiting}} &
Utilizes pre-trained diffusion models with a time-aware encoder for high-quality super-resolution, deblurring, and artifact removal. \\ \cline{2-3}
& \multirow{2}{*}{Real-ESRGAN~\cite{wang2021real}} &
Fast GAN for super-resolution, deblurring, and artifact removal, handling complex real-world degradations efficiently. \\ 
\midrule

\multirow{2}{*}{\textbf{Denoising}} &
\multirow{2.5}{*}{SCUnet~\cite{zhang2023practical}} &
Hybrid UNet-based model combining convolution and transformer blocks, designed for robust denoising under diverse real-world noise conditions. \\ 
% & \multirow{2}{*}{Restormer~\cite{zamir2022restormer}} &
% Transformer-based model for high-quality denoising and efficient restoration. \\ 
\midrule

\multirow{5}{*}{\textbf{Compression artifact removal}} &
\multirow{3}{*}{StableSR-turbo~\cite{wang2024exploiting}} &
Utilizes pre-trained diffusion models with a time-aware encoder for high-quality super-resolution, deblurring, and artifact removal. \\ \cline{2-3}
& \multirow{2}{*}{Real-ESRGAN~\cite{wang2021real}} &
Fast GAN for super-resolution, deblurring, and artifact removal, handling complex real-world degradations efficiently. \\ 
\midrule

\multirow{5}{*}{\textbf{Deblurring}} &
\multirow{3}{*}{StableSR-turbo~\cite{wang2024exploiting}} &
Utilizes pre-trained diffusion models with a time-aware encoder for high-quality super-resolution, deblurring, and artifact removal. \\ \cline{2-3}
& \multirow{2}{*}{Real-ESRGAN~\cite{wang2021real}} &
Fast GAN for super-resolution, deblurring, and artifact removal, handling complex real-world degradations efficiently.. \\ 
\midrule

\multirow{4}{*}{\textbf{Deraining}} &
IDT~\cite{xiao2022image} &
Transformer-based model for de-raining and raindrop removal. \\ \cline{2-3}
& \multirow{1}{*}{UDR-S2Former~\cite{Chen_2023_ICCV}} &
An uncertainty-aware transformer model for rain streak removal.\\ \cline{2-3}
& \multirow{2}{*}{Img2img-turbo-rain~\cite{parmar2024one}} &
Efficient model based on SD-turbo, designed for fast and effective rain removal in real-world images. \\ 
\midrule

\textbf{Raindrop removal} &
IDT~\cite{xiao2022image} &
Transformer-based model for de-raining and raindrop removal. \\ 
\midrule

\multirow{4}{*}{\textbf{Dehazing}} &
\multirow{2}{*}{RIDCP~\cite{wu2023ridcp}} &
Efficient dehazing model utilizing high-quality codebook priors to handle complex real-world haze. \\ \cline{2-3}
& \multirow{2}{*}{KANet~\cite{feng2024advancing}} &
Efficient dehazing network using a localization-and-removal pipeline to handle complex real-world hazy. \\ 
\midrule

\multirow{4}{*}{\textbf{Desnowing}} &
\multirow{2}{*}{Img2img-turbo-snow~\cite{parmar2024one}} &
Efficient model for removing snow artifacts while preserving natural scene details. \\ \cline{2-3} 
& \multirow{2}{*}{Snowformer~\cite{chen2022snowformer}} &
Transormer-based model for removing snowflakes while preserving natural scene details. \\ 
\midrule

\multirow{6}{*}{\textbf{Low-light enhancement}} &
\multirow{2}{*}{Img2img-turbo-night~\cite{parmar2024one}} &
Fast and efficient model based on SD-turbo, designed for low-light enhancement in real-world scenarios. \\ \cline{2-3}
&\multirow{3}{*}{HVI-CIDNet~\cite{yan2024you}} &
Lightweight transformer for low-light and exposure correction, enhancing both image quality and downstream vision tasks efficiently. \\ \cline{2-3}
& \multirow{3}{*}{LightenDiff~\cite{jiang2024lightendiffusion}} &
Diffusion-based framework for low-light enhancement, leveraging Retinex theory and latent-space decomposition for high-quality unsupervised restoration. \\ 
\bottomrule
\end{tabular}
    }
\end{table*}

\subsection{Details of Model Setups}
\textbf{Model Architecture.} In this study, JarvisIR primarily adopts the architecture from Llava-Llama3-8B~\cite{liu2024visual}. Specifically, the input images and instruction texts are first tokenized, then fused, and finally processed by the Large Language Model (LLM) for response generation. (a) Tokenization of input images and instruction texts: We use a frozen CLIP pre-trained ViT-L/14~\cite{radford2021learning} as the image encoder to convert input images into visual tokens. The instruction texts are tokenized into textual tokens using the SentencePiece tokenizer~\cite{kudo2018sentencepiece}. To bridge the different embedding spaces of visual and textual tokens, we implement a trainable image projector to map visual tokens into the textual space, following~\cite{touvron2023llama,zhu2023minigpt}. (b) Token Fusion: We integrate the visual tokens into predefined positions within the textual tokens to achieve token fusion. (c) Response Generation Using LLM: The fused tokens are fed into the LLM to generate the final response. In our experiments, we primarily use Llama3-8B~\cite{touvron2023llama}. Even with their advanced features, pre-trained LLMs lack the ability to furnish accurate responses, thorough reasoning regarding degradation, and precise restoration plans without dataset-specific fine-tuning. Therefore, we employ a full parameter fine-tuning technique that efficiently unleashes the potential of LLM to the maximum extent.

\noindent\textbf{Model setup.} 
Since the CLIP pre-trained ViT-L/14~\cite{radford2021learning} encodes each $14\times 14$ image patch into a visual token, the input image dimensions must be integer multiples of 14. Therefore, we zero-pad the input images to meet this requirement. 
% Although we aim to preserve the original image resolution, constrained by computation resources, if the image is too large ($>672$), we bi-linearly resize the image while maintaining its original ratios. 
We encode the image patches into visual tokens using the CLIP pre-trained ViT-L/14~\cite{radford2021learning}, where each token is a 1024-dimensional vector. These visual tokens are subsequently projected by the image projection layer into the LLM's hidden dimension of 4096.

\noindent\textbf{Training setup.} 
Both the SFT and MRRHF tuning phases utilize the Adam optimizer with learning rate 1e-5 with cosine decay. The warmup ratio is set to 0.03, the maximum sequence length is 2048, and the weight decay is 4. JarvisIR-SFT undergoes training for three epochs with a batch size of 128, while JarvisIR-MRRHF is trained for three epochs using a batch size of 2. During the MRRHF tuning phase, the diverse beam search settings include a size of 3, 5 beam groups, a diversity penalty of 2.0, and a sampling temperature of 0.8. Training is conducted on 8 GPUs (NVIDIA A100 80G).

\begin{figure*}[!t]
    \centering
    \setlength{\belowcaptionskip}{-0.3cm}%调整caption与下文的距离
    \includegraphics[width=1\linewidth]{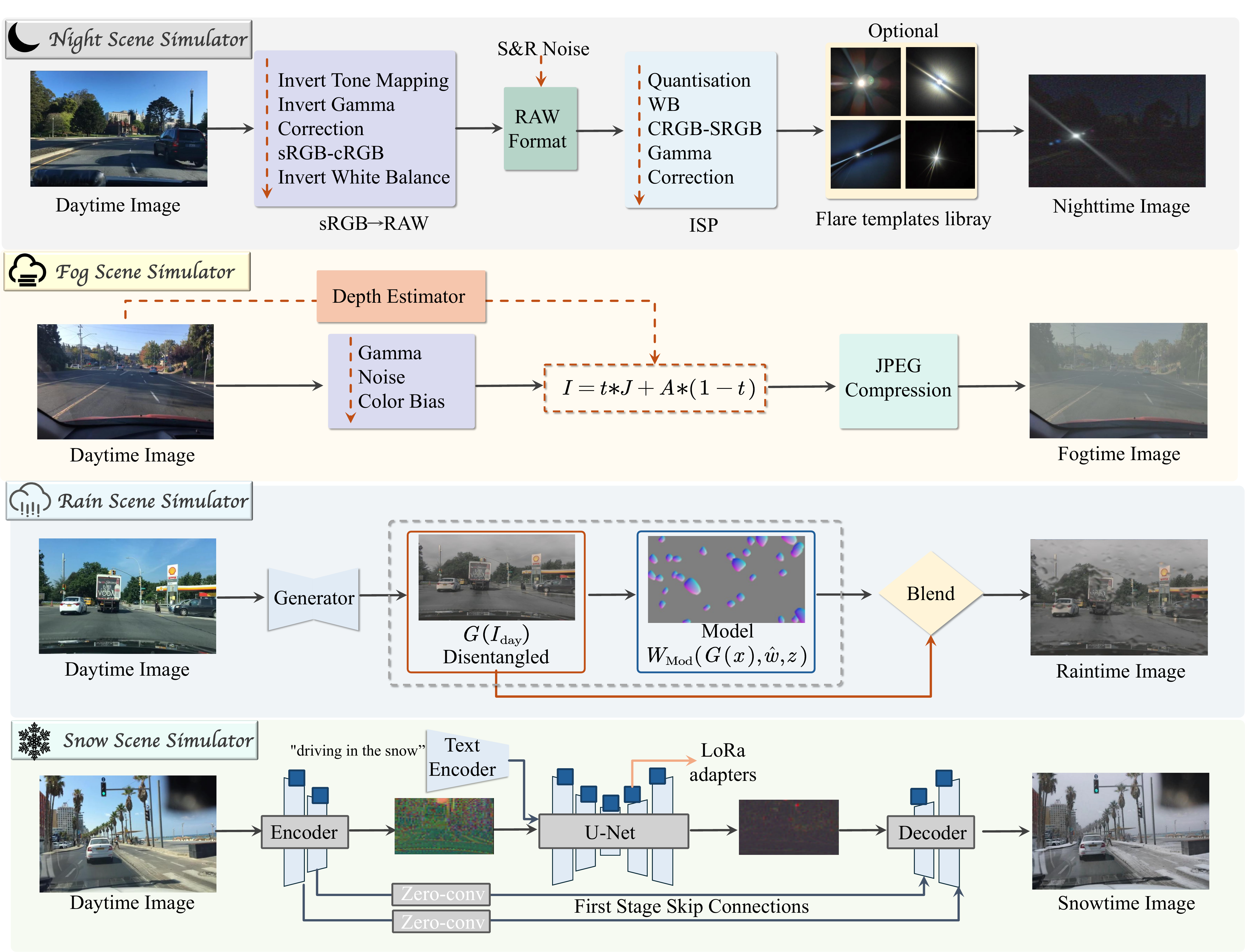}
    \caption{Adverse weather scene simulator. To simulate realistic adverse weather scenarios, including rainy, nighttime, snowy, and foggy, we customized the degradation library developed using physical models and image transformation techniques to synthesize degraded images.}\label{simulators}
\end{figure*}

\section{CleanBench dataset details}\label{Datasets}
\subsection{Dataset statistics}
\textbf{CleanBench.} 
In constructing the CleanBench process, we collected large-scale raw daytime images from various sources, including autonomous driving datasets~\cite{caesar2020nuscenes,caesar2020nuscenes,yu2020bdd100k,sakaridis2021acdc} and natural datasets~\cite{yang2021sparse,cai2018learning,zhou2022lednet,jin2025raindrop,lin2024nighthaze,li2018benchmarking} .The CleanBench dataset contains a total of 150K degraded-clean image pairs. For the construction of CleanBench-Real, we gathered 80K real degraded images consisting of night scenes, fog scenes, snow scenes and rain scenes. These data come from diverse sources, including the aforementioned autonomous driving datasets. Additionally, to enhance the generalizability of JarvisIR in natural contexts, we incorporated natural adverse weather scenes from internet and public datasets~\cite{yang2021sparse,cai2018learning,zhou2022lednet,jin2025raindrop,lin2024nighthaze,li2018benchmarking,liu2018desnownet,quan2021removing}.

\subsection{Details of degradation library}
\begin{figure*}[!t]
    \centering
    \includegraphics[width=1\linewidth]{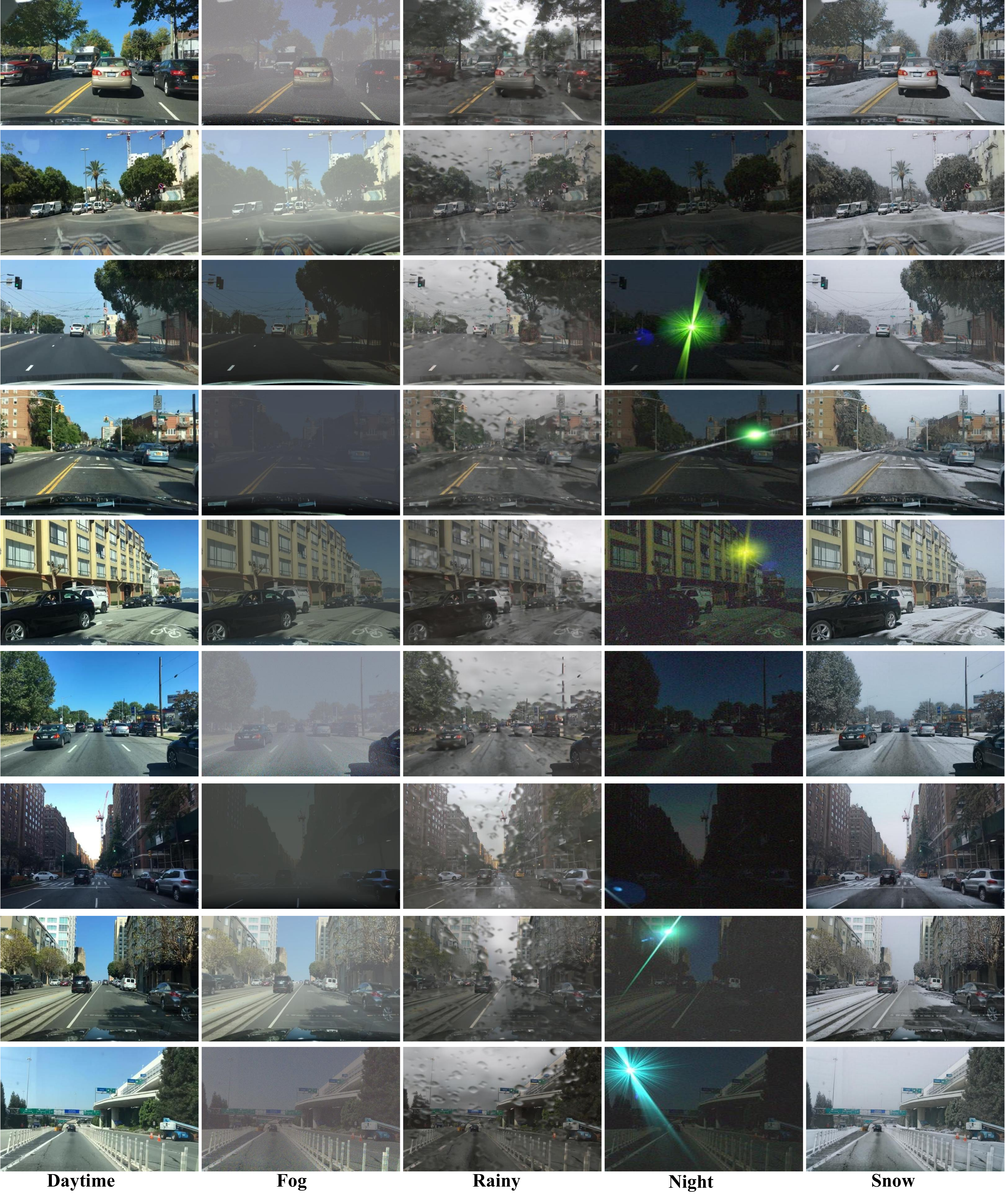}
    \caption{Examples of synthetic adverse weather scenarios in autonomous driving from the CleanBench dataset.}\label{examples_synthetic}
\end{figure*}

As described in Sec 3.1 of the manuscript, we simulate realistic adverse weather scenarios—rainy, nighttime, snowy, and foggy conditions—by customizing a degradation library developed with physical models and image transformation techniques to synthesize degraded images. In this section, we detail our degradation implementations, covering the principles, formulas, and severity setups for the Night Scene Simulator, Fog Scene Simulator, Rain Scene Simulator, and Snow Scene Simulator. Examples for each implementation are provided in Figure~\ref{examples_synthetic}.

\noindent\textbf{Night Scene Simulator.}
Inspired by the work of~\cite{cui2021multitask}, we employ a low-light degradation transform to synthesize realistic low-light images, denoted as $T_{night}$, as illustrated in Figure~\ref{simulators}. Specifically, we first convert the daytime image $I_{day}$ into RAW data using the sRGB→ RAW process~\cite{brooks2019unprocessing}. Next, we linearly attenuate the RAW image and introduce Shot and Read (S\&R) noise, which is commonly observed in camera imaging systems~\cite{brooks2019unprocessing}. Finally, we apply the Image Signal Processing (ISP) pipeline to convert the low-light sensor data back into sRGB format. Additionally, we incorporate flare degradation using flare templates from the Flare7K++~\cite{dai2023flare7k++} dataset. The complete low-light degradation transform $T_{night}$ is given by:
\begin{equation}
T_{\mathrm{night}}\left( \mathrm{I}_{\mathrm{day}} \right) =T_{ISP}\left( T_{sRGB\rightarrow RAW}\left( \mathrm{I}_{\mathrm{day}} \right) +\mathrm{I}_{\mathrm{noise}} \right) +\mathrm{I}_{\mathrm{flare}},
    \label{night_simulator}
\end{equation}
which generates a degraded image $I_{day}$ that closely resembles a dark nighttime scene. Furthermore, we use an online dynamic degradation process. It applies randomized parameter combinations, as defined in Equation~\ref{night_simulator}, to simulate diverse nighttime driving conditions.

\noindent\textbf{Fog Scene Simulator.}
Inspired by RIDCP~\cite{wu2023ridcp}, we design a foggy image degradation transform, denoted as $T_{\text {fog }}$, to synthesize realistic hazy images, as shown in Figure~\ref{simulators}. Specifically, we simulate fog by introducing transmission maps $t(x)$ using depth estimation algorithms (e.g., Depth anything V2~\cite{yang2024depth}), combined with exponential attenuation $e^{\beta d(x)}$, where $\beta$ controls haze density within the range $[0.3,1.5]$. Additionally, poor lighting conditions are modeled by applying a brightness adjustment factor $\gamma \in$ $[1.5,3.0]$, Gaussian noise $\mathcal{N}$, and atmospheric light variation $A+\Delta A$, where $\Delta A$ is sampled from $[-0.025,0.025]$. To further enhance realism, JPEG compression artifacts are introduced by applying JPEG $(\cdot)$ to the degraded image. The complete foggy image synthesis process is defined as:
\begin{equation}
T_{\text {fog }}\left(I_{\text {day }}\right)=\operatorname{JPEG}\left(\mathcal{P}\left(I_{\text {day }}^\gamma+\mathcal{N}, e^{\beta d(x)}, A+\Delta A\right)\right),
\end{equation}
where $\mathcal{P}$ represents the hazy image formation process, $I_{\text {day}}$ is the clean image, and $d(x)$ is the estimated depth map. The variable $x$ refers to the spatial coordinates of the image. This dynamic degradation process is designed to operate online with randomized parameters, simulating diverse real-world foggy conditions.

\noindent\textbf{Rain Scene Simulator.}
Inspired by PGDGN~\cite{pizzati2023physics}, we introduce a rain degradation transform, denoted as $T_{\text{rain}}$, to generate realistic rainy images (Figure~\ref{simulators}). This transform synthesizes rainy images by combining a disentangled clean image with a physics-based rain rendering model. The degradation process is formulated as:
\begin{equation}
T_{\mathrm{rain}}\left( I_{\mathrm{day}} \right) =W_{\mathrm{Mod}}(G(I_{\mathrm{day}}),\hat{w},z),
\end{equation}
where $I_{\text {day}}$ is the clean image, $G(I_{\mathrm{day}})$ represents the disentangled base image, and $W_{\text {Mod}}$ is the rain rendering model. $W_{\text {Mod}}$ incorporates parameters $\hat{w}=\left\{\hat{w}_d, \hat{w}_{n d}\right\}$, with $\hat{w}_d$ controlling differentiable aspects such as raindrop size and streak density, and $\hat{w}_{n d}$ addressing nondifferentiable properties. The term $z$ introduces stochastic noise for variability in rain effects. This process applies $W_{\text {Mod}}$ to add realistic raindrop occlusions, rain streaks, and scene wetness to the disentangled image. $G(I_{\mathrm{day}})$, generating a visually plausible rainy image $T_{\text {rain}}\left(I_{\text {day}}\right)$ with controlled and diverse effects.

\noindent\textbf{Snow Scene Simulator.}
Building on the img2img-turbo model~\cite{parmar2024one}, we introduce a snow transformation, denoted as $T_{\text{snow}}$, to generate realistic snowy images from daytime inputs. This process uses the SD-Turbo model with textual conditioning. It synthesizes snowy scenes by combining the input image with a latent diffusion-based generator and a textual prompt. The snow transformation is formulated as:
\begin{equation}
    T_{\text {snow }}\left(I_{\text {day }}, C_{\text {snow }}\right)=G_{\text {snow }}\left(I_{\text {day }}, C_{\text {snow }}\right),
\end{equation}
where $I_{\text{day}}$ is the daytime input image, $C_{\text{snow}}$ is the textual condition (e.g., ``driving in the heavy snow"), and $G_{\text{snow}}$ represents the generator. By employing LoRA adapters and skip connections, the generator enables precise control over scene characteristics while maintaining the structural integrity of the input image. This process applies $G_{\text{snow}}$ to infuse the daytime image $I_{\text{day}}$ with snowy features, guided by the contextual information in $C_{\text{snow}}$. The resulting synthetic image aligns closely with the visual expectations of a snowy environment while maintaining consistency with the original scene's structure.

\begin{table*}[t]
\centering
\setlength{\abovecaptionskip}{0.1cm} % 调整caption与图的距离
\setlength{\belowcaptionskip}{-0.1cm} % 调整caption与下文的距离
\caption{Ablation studies on tuning paradigm, differentiated contrast weights, different sample generation strategies, and entropy regularization. The ``Reward" represents the average reward scores obtained during alignment tuning, spanning from -1 to 1. A negative score indicates a penalty, while a positive score represents a reward. The ``Diversity" reflects the average number of unique responses produced during the training process. Additionally, we evaluate performance on the CleanBench-Real validation set using four non-reference metrics: MUSIQ, MANIQA, CLIP-IQA+, and LIQE. The reported values represent the average performance across all tested scenes.}\label{strategy2}
\scalebox{0.97}{
\setlength\tabcolsep{6pt}
\renewcommand\arraystretch{1}
\begin{tabular}{ccccccc}
\toprule
Strategy & Reward & Diversity & MUSIQ $\uparrow$ & MANIQA $\uparrow$ & CLIP-IQA+ $\uparrow$ & LIQE $\uparrow$ \\
\midrule
Vanilla RRHF & 0.40 & 3.12 & 63.89 & 0.5090 & 0.5388 & 3.589 \\
\rowcolor{yellow!20} MRRHF (\textbf{Ours}) & \textbf{0.67} & \textbf{6.55} & \textbf{71.43} & \textbf{0.7099} & \textbf{0.7296} & \textbf{4.411} \\
\midrule
w/o. differentiated contrast weights & 0.53 & 2.62 & 63.22 & 0.5871 & 0.6130 & 3.597 \\
\rowcolor{yellow!20} w. differentiated contrast weights (\textbf{Ours}) & \textbf{0.67} & \textbf{6.55} & \textbf{71.43} & \textbf{0.7099} & \textbf{0.7296} & \textbf{4.411} \\
\midrule
offline sample generation & 0.43 & 3.63 & 64.12 & 0.5323 & 0.6012 & 3.620 \\
online sample generation & -0.87 & 1.27 & - & - & - & - \\
\rowcolor{yellow!20} hybrid sample generation (\textbf{Ours}) & \textbf{0.67} & \textbf{6.55} & \textbf{71.43} & \textbf{0.7099} & \textbf{0.7296} & \textbf{4.411} \\
\midrule
w/o. entropy regularization & 0.50 & 4.56 & 65.06 & 0.6207 & 0.6915 & 3.867 \\
\rowcolor{yellow!20} w. entropy regularization (\textbf{Ours}) & \textbf{0.67} & \textbf{6.55} & \textbf{71.43} & \textbf{0.7099} & \textbf{0.7296} & \textbf{4.411} \\

\bottomrule
\end{tabular}}
\vspace{-0.3cm}
\end{table*}

\section{More ablation}\label{ablation}
To thoroughly investigate the proposed JarvisIR, we conducted an extensive array of ablation studies on the CleanBench-Real dataset. Four non-reference metrics are used for assessment:  MUSIQ~\cite{ke2021musiq}, MANIQA~\cite{yang2022maniqa}, CLIP-IQA+~\cite{wang2023exploring}, LIQE~\cite{zhang2023blind}. The specific elements of these studies are further expounded in the sections that follow.

\subsection{MRRHF vs. vanilla RRHF} We evaluate the effectiveness of our proposed MRRHF by comparing it with vanilla RRHF~\cite{yuan2023rrhf}. The reward and diversity metrics over training iterations are illustrated in Table~\ref{strategy2}. Fine-tuning JarvisIR with MRRHF significantly improves the average values of both reward and diversity by 0.19 and 3.43, respectively, compared to using RRHF. The degradation in diversity and reward when using vanilla RRHF results from its offline sample generation strategy. As discussed in Sec. 5 in the manuscript, this strategy confines its generated samples to the finite sample space created by the SFT model using diverse beam search~\cite{vijayakumar2016diverse}. In contrast, our MRRHF employs a hybrid sample generation strategy and entropy regularization, providing sufficient sample exploration space to achieve globally optimal results.

\subsection{Sample generation strategy and entropy regularization}
In our manuscript, we examine the effects of the sample generation strategy and entropy regularization on the MRRHF tuning process, focusing on reward scores and response diversity. This section provides further evidence of the effectiveness of our hybrid sample generation strategy and entropy regularization. Specifically, as shown in Table~\ref{strategy2}, we assess their impact on performance using the CleanBench-Real validation set. The results demonstrate that our hybrid sampling approach and entropy regularization not only enhance training stability and facilitate high-quality exploration of the optimization space but also significantly improve testing performance.

\subsection{Effectiveness of differentiated contrast weights} 
In Equation 4 of our manuscript, we refine the original ranking loss~\cite{yuan2023rrhf} by introducing differentiated contrast weights, expressed as $L_{\mathrm{rank}}=\sum_{s_i<s_j}{{\color{red} \left( s_j-s_i \right) }}\max \left( 0,p_i-p_j \right) 
$. The term $\left( s_j-s_i \right)$ represents the differentiated contrast weights. We compare this with the original ranking loss $\hat{L}_{\mathrm{rank}}=\sum_{s_i<s_j}{\max \left( 0,p_i-p_j \right)}$. Table~\ref{strategy2} presents the reward and diversity metrics over training iterations. When the ranking loss is applied without differentiated contrast weights $\hat{L}_{\mathrm{rank}}$ the average values of both reward and diversity decrease by 0.14 and 3.93, respectively, compared to using $L_{\mathrm{rank}}$. We attribute this to the differentiated contrast weights enabling the VLM to recognize that some negative examples are neutral (with reward scores close to positive examples) and thus should not be excessively penalized, which helps prevent confusion during VLM training. Specifically, assuming the system uses diverse beam search to obtain multiple responses $r_1, \ldots, r_i, r_k ,r_{n}$ the original RRHF algorithm treats the best response $r_k$ as positive and the remaining responses $r_i<r_k$ as negative examples of $r_k$ and applies the same penalty to them. However, this approach may not be reasonable, especially when the preference scores of different $r_i$ are similar. For instance, when the preference of $r_{k+1}$ is only slightly worse than $r_k$, while $r_n$ is significantly worse than $r_k$, the model should differentiate and apply different penalty strengths, slightly penalizing $r_{k+1}$ and heavily penalizing $r_n$ compared to $r_k$. To address this, we propose using the score $\mathcal{S}\left(r_i\right)$ from a reward model $\mathcal{S} \left( \cdot \right)$ to indicate the numerical preference of $r_i$, i.e., the differentiated contrast weights $\left( s_j-s_i \right)$.

\subsection{Impact of reasoning for decision-making} 
As the pioneering work~\cite{wei2022chain} points out, Chain-of-Thought (CoT) is ``a series of intermediate reasoning steps" that has proven effective in complex reasoning tasks~\cite{wei2022chain,kojima2022large,zhang2022automatic}. The main idea of CoT is to prompt large language models (LLMs) to output not only the final answer but also the reasoning process leading to it, resembling human cognitive processes. Inspired by this approach, we enable JarvisIR to provide detailed degradation and reasoning insights about the degraded image before making decisions, specifically before producing the task sequence with model selection. To assess the impact of reasoning on final decision-making, we perform ablation experiments on the CleanBench-Real validation set by comparing two variants: (1) directly requesting JarvisIR to output the task sequences, and (2) providing detailed degradation and reasoning insights before outputting the task sequences. As shown in Table~\ref{reasoning}, providing detailed degradation and reasoning insights significantly enhances JarvisIR's decision-making, leading to notable improvements in the four non-reference metrics. By explicitly describing degradations and reasoning insights, the model can use in-context learning to align selected tasks and restoration experts with the specific degradations present. This strategy not only enhances interpretability but also introduces constraints that make the model's decisions more reliable in real-world scenarios.

\begin{table}[t]
\centering
\setlength{\abovecaptionskip}{0.1cm} %调整caption与图的距离
\caption{Ablation studies on the impact of reasoning for decision-making. We evaluate performance on the CleanBench-Real validation set using four non-reference metrics: MUSIQ, MANIQA, CLIP-IQA+, and LIQE. The reported values represent the average performance across all tested scenes.}\label{reasoning}
\scalebox{0.86}{
\setlength\tabcolsep{1pt}
\renewcommand\arraystretch{1}
\begin{tabular}{lcccc}
\toprule 

Configurations & MUSIQ $\uparrow$ & MANIQA $\uparrow$ & CLIP-IQA+ $\uparrow$ & LIQE $\uparrow$ \\
\midrule
w/o. reasoning& 71.17 & 0.6942 & 0.7156 & 4.394 \\
\rowcolor{yellow!20}\textbf{(Ours)} w reasoning & \textbf{71.43} &  \textbf{0.7099} & \textbf{0.7296} & \textbf{4.411} \\

\bottomrule
\end{tabular}}
\vspace{-0.5cm}
\end{table}

\begin{table*}[t]
\centering
\setlength{\abovecaptionskip}{0.1cm} %调整caption与图的距离
\setlength{\belowcaptionskip}{-0.1cm}%调整caption与下文的距离
\caption{Ablation studies on the impact of different reward model configurations. We evaluate performance on the CleanBench-Real validation set using four non-reference metrics: MUSIQ, MANIQA, CLIP-IQA+, and LIQE. The reported values represent the average performance across all tested scenes.}\label{reward_table}
\scalebox{0.93}{
\setlength\tabcolsep{12pt}
\renewcommand\arraystretch{1}
\begin{tabular}{lcccc}
\toprule
Configurations & MUSIQ $\uparrow$ & MANIQA $\uparrow$ & CLIP-IQA+ $\uparrow$ & LIQE $\uparrow$ \\
\midrule
(I) Q-align~\cite{wu2023q} + Q-Instruct~\cite{wu2024q} & 71.41 & 0.7094 & 0.7308 & 4.419 \\
\midrule
(II) Q-align & 71.35 & 0.7086 & 0.7288 & 4.409 \\
(II) Q-Instruct~\cite{wu2024q} & 71.37 & 0.7093 & 0.7257 & 4.402 \\
(II) MUSIQ~\cite{ke2021musiq} & 71.64 & 0.6932 & 0.6977 & 3.955 \\
(II) MANIQA~\cite{yang2022maniqa} & 68.49 & 0.7126 & 0.6805 & 3.981 \\
\midrule
(III) MUSIQ~\cite{ke2021musiq} + MANIQA~\cite{yang2022maniqa} & 71.52 & 0.7118 & 0.7068 & 4.127 \\
\midrule
\rowcolor{yellow!20}(\textbf{Ours}) Q-Instruct~\cite{wu2024q}+ MUSIQ~\cite{ke2021musiq} + MANIQA~\cite{yang2022maniqa}  & \textbf{71.43} &  \textbf{0.7099} & \textbf{0.7296} & \textbf{4.411}\\
\bottomrule
\end{tabular}}
\end{table*}

\subsection{Impact of reward model} 
To analyze how various reward model configurations affect model optimization, we conducted an ablation experiment exploring three distinct settings: (I) multiple VLM-based IQA models as a unifined reward model (e.g., Q-instruct~\cite{wu2024q} and Q-align~\cite{wu2023q}). (II) using a single VLM-based IQA model (e.g., Q-Instruct~\cite{wu2024q} or Q-align~\cite{wu2023q}) or a traditional IQA model (e.g., MUSIQ~\cite{ke2021musiq} or MANIQA~\cite{yang2022maniqa}). (III) multiple traditional IQA models as a unifined model (e.g., MUSIQ~\cite{ke2021musiq} and MANIQA~\cite{yang2022maniqa}). The results of JarvisIR-MRRHF trained with different reward models are summarized in Table~\ref{reward_table}. Based on the results, we make the following observations:
(1) Using multiple VLM-based IQA models as the reward model significantly improves perception metrics, although it increases resource consumption during training.
(2) Training with a single IQA model improves the corresponding metric significantly, but other metrics may experience some degradation.
(3) Combining multiple traditional IQA models as the reward model enhances performance on certain metrics, but the improvements are asymmetrical—some traditional metrics exhibit very high performance while perception metrics are relatively low. Consequently, we opt to create the unified reward model by combining both VLM-based and non-VLM-based IQA models, such as Q-instruct~\cite{wu2024q}, MUSIQ~\cite{ke2021musiq}, and MANIQA~\cite{yang2022maniqa}. This combination allows for a comprehensive evaluation of system responses while preserving training efficiency.

\begin{figure*}[!t]
    \centering
\setlength{\abovecaptionskip}{-0.1cm} %调整caption与图的距离
    \setlength{\belowcaptionskip}{-0.3cm}%调整caption与下文的距离
    \includegraphics[width=1\linewidth]{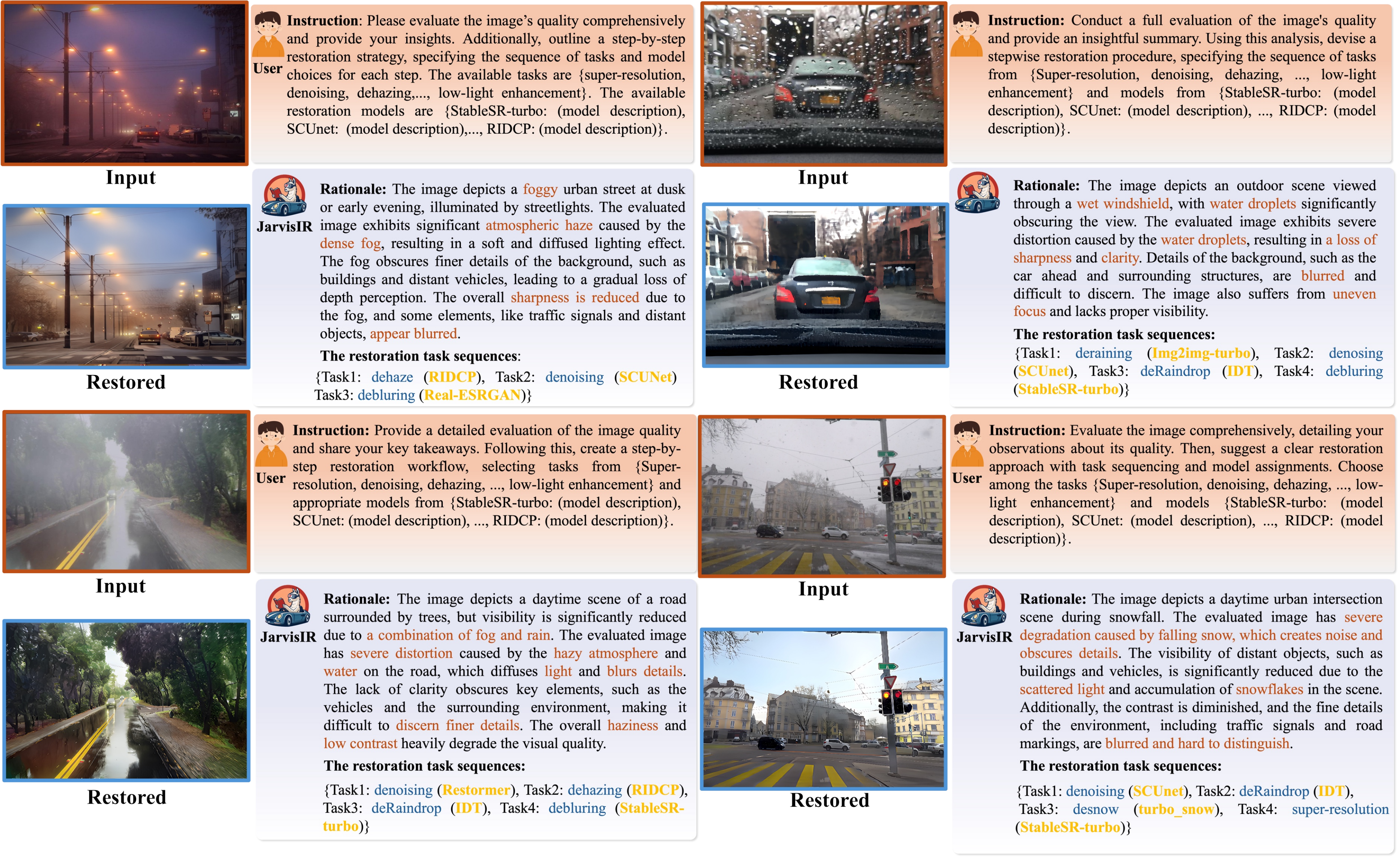}
    \caption{More examples of JarvisIR's perception restoration are presented. Initially, JarvisIR assesses the degradation of the input images and parses user instructions to formulate a task plan, selecting appropriate expert models for each subtask. The selected experts perform their designated tasks and return the results to JarvisIR, which integrates the outcomes and provides the final answer to the user.}\label{resoning}
\end{figure*}

\begin{figure*}[!t]
    \centering
    \setlength{\belowcaptionskip}{-0.5cm}%调整caption与下文的距离
    \includegraphics[width=1\linewidth]{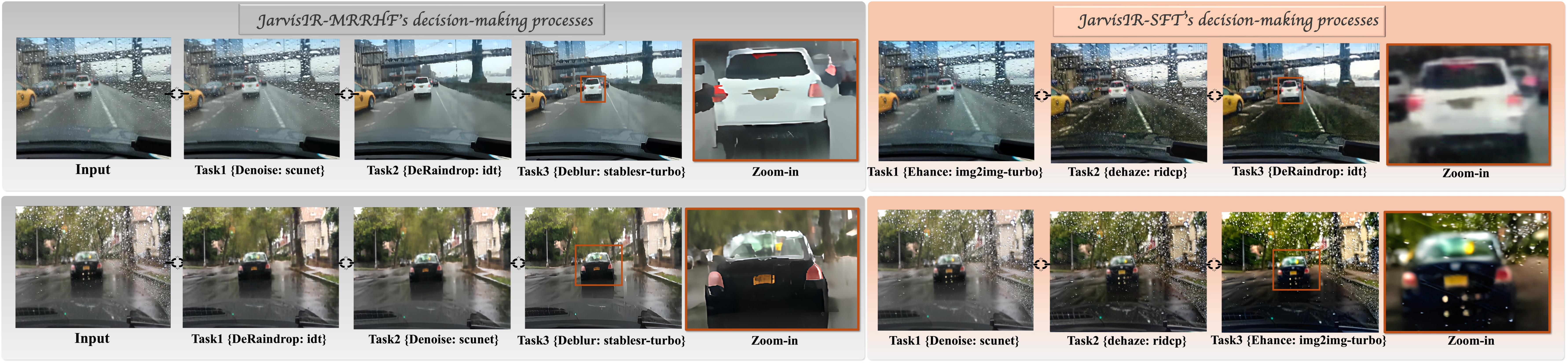}
    \caption{Comparison of the decision-making processes of JarvisIR-MRRHF and JarvisIR-SFT. The results indicate that the MRRHF version accurately predicts the correct task sequence and selects appropriate restoration models. Conversely, the SFT version often fails to make suitable decisions in real-world scenarios due to the domain gap between training and real data distributions.}\label{decision_make}
\end{figure*}

\section{More visual results.}\label{Results}
\subsection{Perception restoration}
Additional visual comparisons highlight the effectiveness of the proposed JarvisIR framework in real-world adverse weather conditions. Figure~\ref{resoning} illustrates the comprehensive workflow of JarvisIR, which begins by receiving user commands and degraded images. JarvisIR evaluates the image quality, identifies degradation factors, and formulates task sequences. It then selects appropriate models for tasks such as denoising, dehazing, and super-resolution. The outputs include evaluated inference insights, detailed restoration plans, and enhanced images, effectively bridging user instructions with image restoration plans. 

Figure~\ref{decision_make} illustrates the decision-making processes of both JarvisIR-MRRHF and JarvisIR-SFT. Experimental results indicate that the decision-making capability of JarvisIR-MRRHF surpasses that of JarvisIR-SFT. Specifically, JarvisIR-MRRHF makes correct decisions in cases where JarvisIR-SFT previously failed. For example, in coupled degraded real rain scenarios (the first row), JarvisIR-SFT yields a mediocre decision—``Enhancement (Img2img-turbo) → Dehaze (RIDCP) → DeRaindrop (IDT)"—which does not remove raindrops and blur the background. However, JarvisIR-MRRHF accurately identifies the appropriate restoration tasks and selects the optimal models to solve them: ``Denoise (SCUNet) → DeRaindrop (IDT) → Deblur (StableSR-turbo)". This improvement confirms that MRRHF fine-tuning significantly enhances JarvisIR's decision-making ability under real-world conditions, reduces hallucination errors, and improves generalization performance.

\begin{figure*}[!t]
    \centering
    \includegraphics[width=1\linewidth]{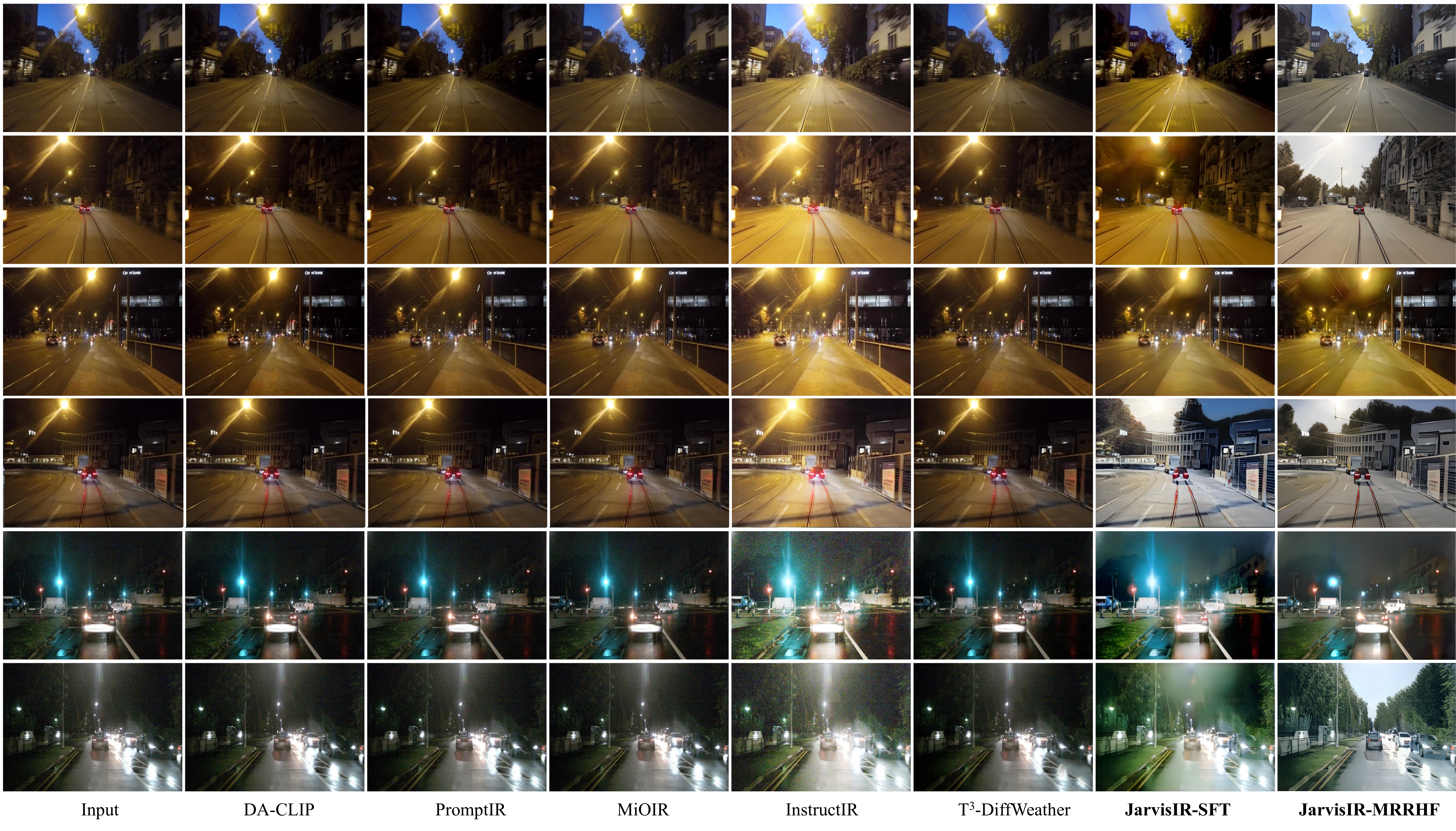}
    \caption{Visual comparisons among various methods on CleanBench-Real's night scene validation set.}\label{night_supple}
\end{figure*}

\begin{figure*}[!t]
    \centering
    \includegraphics[width=1\linewidth]{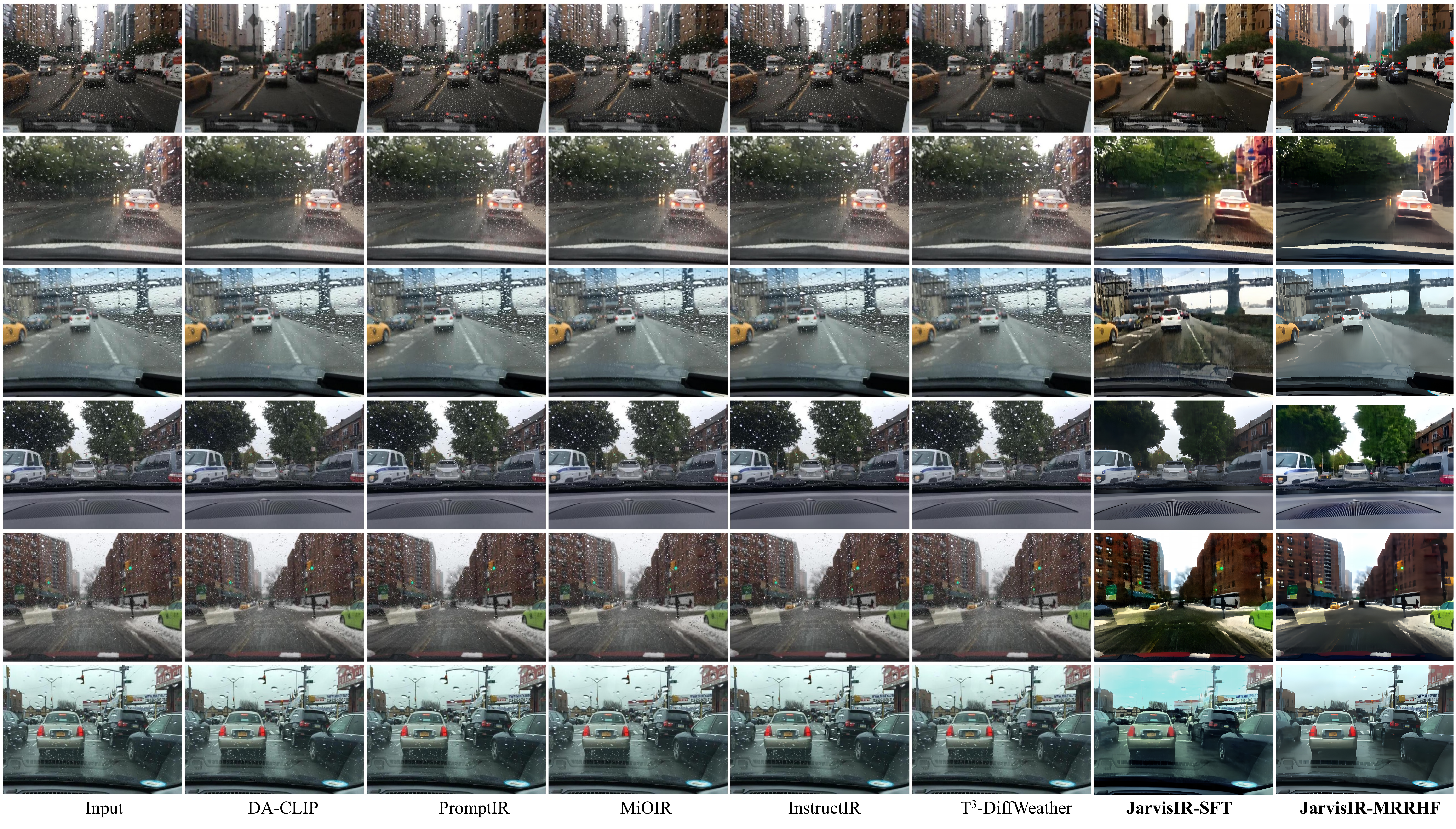}
    \caption{Visual comparisons among various methods on CleanBench-Real's rain scene validation set.}\label{rain_supple}
\end{figure*}

\begin{figure*}[!t]
    \centering
    \includegraphics[width=1\linewidth]{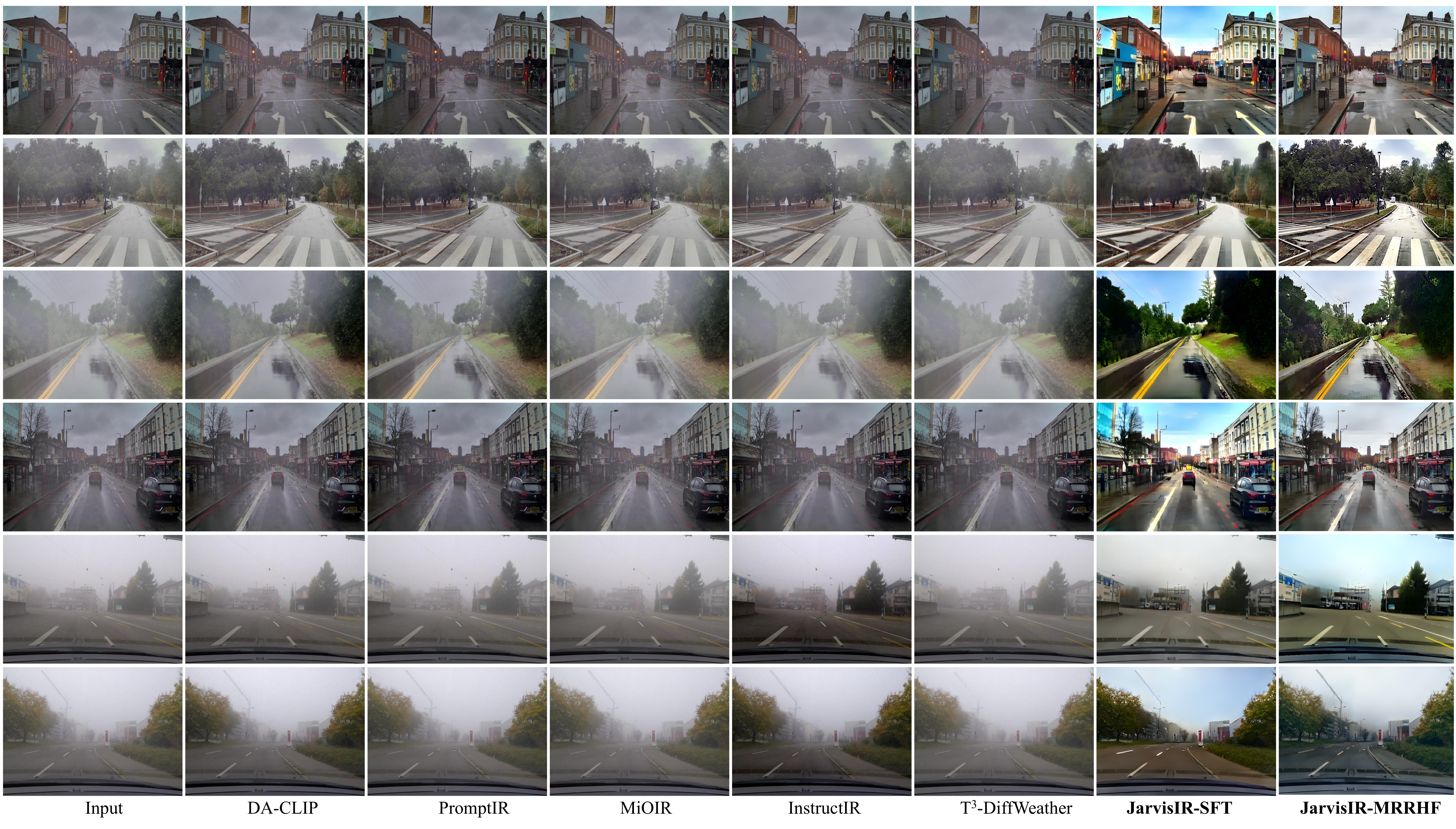}
    \caption{Visual comparisons among various methods on CleanBench-Real's fog scene validation set.}\label{fog_supple}
\end{figure*}

\begin{figure*}[!t]
    \centering
    \includegraphics[width=1\linewidth]{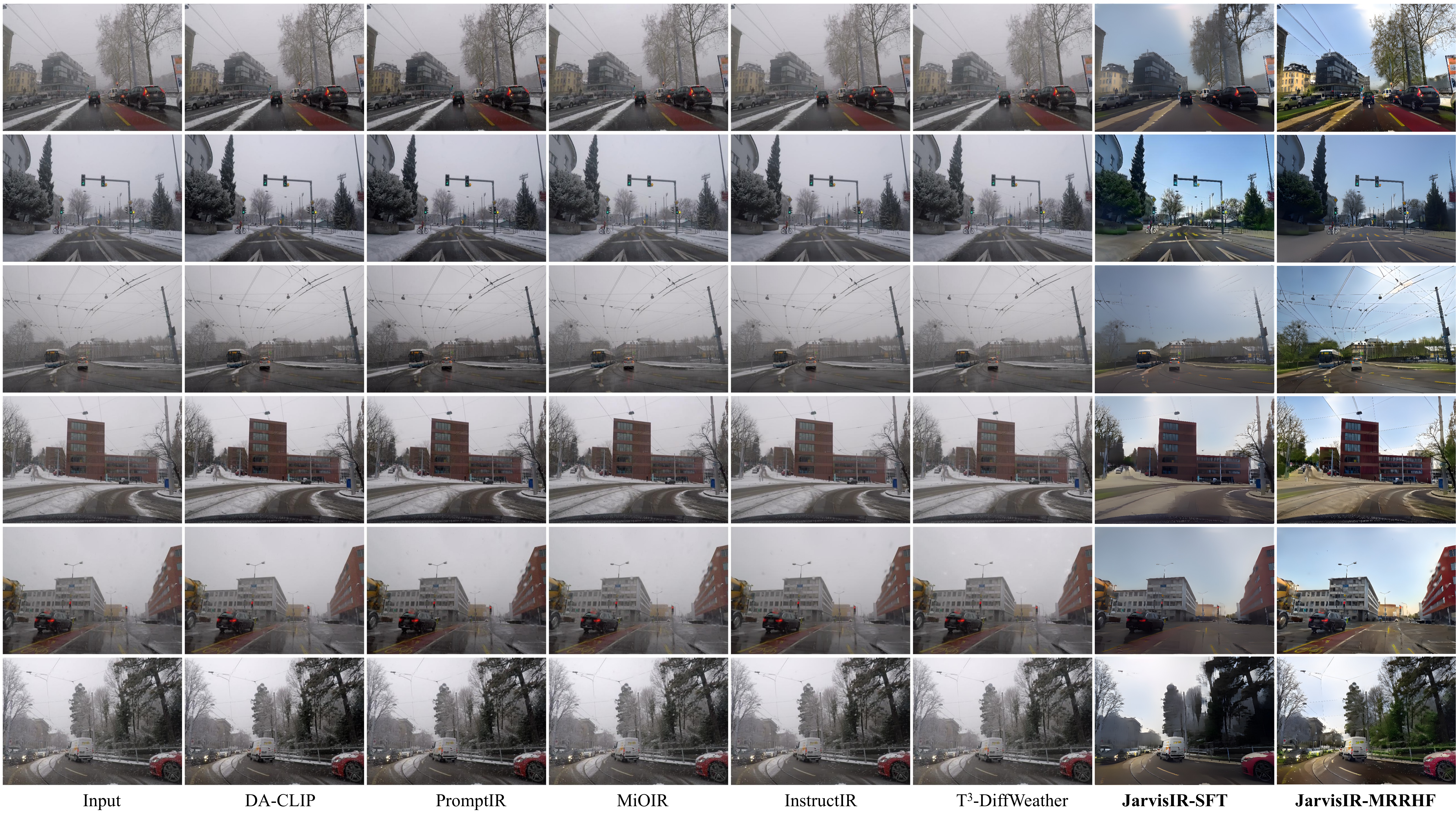}
    \caption{Visual comparisons among various methods on CleanBench-Real's snow scene validation set.}\label{snow_supple}
\end{figure*}

Figures \ref{night_supple}, \ref{rain_supple}, \ref{fog_supple}, and \ref{snow_supple} illustrate visual comparisons of our method and the baseline methods across four different scenes on the CleanBench-Real test set. Our results demonstrate that JarvisIR outperforms the comparative methods in terms of color enhancement, detail preservation, and the elimination of degradations, achieving a superior balance among these aspects.
Conversely, the baseline methods perform poorly in real-world environments. They struggle to handle coupled degradations that occur simultaneously in natural settings, such as low light combined with fog or a mixture of rain and fog. These limitations may arise from their heavy dependence on specific degradation priors and significant domain gaps due to mismatches between synthetic training data distributions and real-world data. Consequently, they often produce subpar recovery results featuring artifacts, overexposure, underexposure, and amplified noise.

\section{Limitations, broader impacts and future work}\label{Future}
The primary limitation of our research is that JarvisIR is unable to address all real-world restoration scenarios. While it demonstrates effectiveness in handling most degradation scenarios relevant to autonomous driving, it does not extend to tasks such as underwater image restoration, old photo enhancement, or blind face restoration. By incorporating appropriate data and tools, rapid adaptation could be achieved through the proposed training paradigm. Furthermore, the tools currently employed are limited in scope and capability. In our future work, we will incorporate more advanced and robust restoration tools that might further enhance JarvisIR’s ability to address real-world coupled degradation challenges.

Another future work could focus on retaining the original image resolution during training. Most current vision-language models (VLMs) resize input images to a fixed resolution, such as $336 \times 336$, which may degrade performance, as resolution variation may affect the model's perception of degradation. To mitigate this, future research could explore techniques to maintain original image resolutions. One approach involves adapting the position embeddings in CLIP~\cite{radford2021learning} using bicubic interpolation to accommodate varying image dimensions.

This work focuses on building an autonomous, robust, intelligent restoration system tailored for real-world challenges. To enhance system robustness, reduce hallucinations, and improve generalizability, we introduce a novel two-stage framework that integrates supervised fine-tuning with human feedback alignment. By utilizing human feedback and large-scale real unlabeled data, our method allows the VLM to be fine-tuned in an unsupervised manner. We believe that this paradigm can inspire future work to build more powerful and versatile intelligent systems.

\begin{table*}[t!]
\centering
\caption{Instruction generated by GPT-4V using the self-instruct strategy~\cite{wang2022self}}
\begin{tabular}{p{1cm} p{14cm}}
\toprule 
\textbf{\#} & \textbf{Instruction} \\ \hline
1 & Please evaluate the image’s quality comprehensively and provide your insights. Additionally, outline a step-by-step restoration strategy, specifying the sequence of tasks and model choices for each step. The available tasks are {super-resolution, denoising, dehazing,..., low-light enhancement}. The available restoration models are {StableSR-turbo: (model description), SCUnet: (model description),..., RIDCP: (model description)}. \\ \midrule
2 & Analyze the quality of the image comprehensively and provide your insights. Furthermore, propose a restoration strategy by detailing each task and model choice sequentially. The available tasks include {super-resolution, denoising, dehazing,..., low-light enhancement}. The restoration models are {StableSR-turbo: (model description), SCUnet: (model description),..., RIDCP: (model description)}. \\ \midrule
3 & Assess the overall quality of the image and provide a detailed evaluation. Then, design a step-by-step restoration process, specifying tasks and model choices. Tasks available are {super-resolution, denoising, dehazing,..., low-light enhancement}, and models include {StableSR-turbo: (model description), SCUnet: (model description),..., RIDCP: (model description)}. \\ \midrule
4 & Perform a comprehensive evaluation of the image quality and explain your observations. Additionally, develop a step-by-step restoration plan, identifying tasks and model choices. Available tasks are {super-resolution, denoising, dehazing,..., low-light enhancement}, and models include {StableSR-turbo: (model description), SCUnet: (model description),..., RIDCP: (model description)}. \\ \midrule
5 & Conduct a thorough analysis of the image’s quality and provide your insights. Subsequently, create a restoration strategy step by step, specifying the tasks and model choices. The tasks available are {super-resolution, denoising, dehazing,..., low-light enhancement}, and models are {StableSR-turbo: (model description), SCUnet: (model description),..., RIDCP: (model description)}. \\ \midrule
6 & Evaluate the quality of the image comprehensively and outline your findings. Moreover, formulate a sequential restoration plan, detailing tasks and model selections. Available tasks include {super-resolution, denoising, dehazing,..., low-light enhancement}, and models are {StableSR-turbo: (model description), SCUnet: (model description),..., RIDCP: (model description)}. \\ \midrule
7 & Provide a detailed assessment of the image’s quality and share your observations. Then, create a restoration strategy in a step-by-step manner, specifying tasks and models. Available tasks are {super-resolution, denoising, dehazing,..., low-light enhancement}, and models include {StableSR-turbo: (model description), SCUnet: (model description),..., RIDCP: (model description)}. \\ \midrule
8 & Examine the image comprehensively and describe its quality in detail. Additionally, outline a stepwise restoration process, specifying the tasks and models for each stage. Tasks available are {super-resolution, denoising, dehazing,..., low-light enhancement}, and models are {StableSR-turbo: (model description), SCUnet: (model description),..., RIDCP: (model description)}. \\ \midrule
9 & Review the image quality comprehensively and present your findings. Furthermore, propose a restoration plan step by step, detailing the tasks and models used. Available tasks are {super-resolution, denoising, dehazing,..., low-light enhancement}, and restoration models are {StableSR-turbo: (model description), SCUnet: (model description),..., RIDCP: (model description)}. \\ \midrule
10 & Perform a detailed evaluation of the image and provide your insights. Then, construct a step-by-step restoration strategy, specifying the sequence of tasks and the corresponding models. The available tasks include {super-resolution, denoising, dehazing,..., low-light enhancement}, and models are {StableSR-turbo: (model description), SCUnet: (model description),..., RIDCP: (model description)}. \\ 
\bottomrule 
\end{tabular}
\label{table:instruction_pool}
\end{table*}

\begin{table*}[t!]
\centering
\caption{Responses generated by GPT-4V using the self-instruct strategy~\cite{wang2022self}}
\begin{tabular}{p{1cm} p{14cm}}
\toprule 
\textbf{\#} & \textbf{Response} \\ \hline
1 & The image depicts an outdoor scene viewed through a wet windshield, with water droplets significantly obscuring the view. The evaluated image exhibits severe distortion caused by the water droplets, resulting in a loss of sharpness and clarity. Details of the background, such as the car ahead and surrounding structures, are blurred and difficult to discern. The image also suffers from uneven focus and lacks proper visibility. \\ \midrule
2 & This image portrays an outdoor scene as seen through a wet windshield, where water droplets heavily obstruct the view. The distortion caused by these droplets leads to a noticeable reduction in sharpness and clarity. Background elements, like the car in front and nearby structures, appear heavily blurred and challenging to distinguish. The image further suffers from uneven focus, reducing overall visibility. \\ \midrule
3 & An outdoor setting is captured in this image, viewed through a wet windshield. The water droplets significantly obscure the scene, causing severe distortion. Sharpness and clarity are greatly reduced, and background features, such as the car ahead and surrounding buildings, are blurred and hard to recognize. Uneven focus further detracts from the image’s overall visibility. \\ \midrule
4 & This image showcases an outdoor scene as observed through a wet windshield, with water droplets causing significant obstruction. The distortion leads to substantial clarity and sharpness loss. Details in the background, including the car ahead and nearby structures, are blurred and hard to identify. Additionally, the image has uneven focus, which impairs its overall visibility. \\ \midrule
5 & The image illustrates an outdoor view observed through a wet windshield, where water droplets obscure much of the scene. The distortion caused by the droplets results in a notable reduction in sharpness and clarity. Background details, such as the car ahead and surrounding structures, appear blurry and indistinct. Uneven focus further degrades the visibility of the image. \\ \midrule
6 & In this image, an outdoor setting is depicted as seen through a wet windshield, with water droplets heavily obscuring the view. The distortion from the droplets results in a considerable loss of sharpness and clarity. Background elements, like the car ahead and nearby buildings, are blurred and hard to discern. The uneven focus of the image adds to its overall lack of visibility. \\ \midrule
7 & This image shows an outdoor scene viewed through a wet windshield, with water droplets creating significant visual obstruction. The distortion caused by the droplets reduces the sharpness and clarity of the image. Background features, including the car ahead and the surrounding structures, are blurred and difficult to distinguish. The image also suffers from uneven focus, limiting its visibility. \\ \midrule
8 & An outdoor view is displayed in this image, captured through a wet windshield with water droplets blocking much of the scene. The distortion caused by the droplets results in reduced clarity and sharpness. Background details, such as the car in front and nearby structures, are blurred and indistinct. Uneven focus further compromises the visibility of the image. \\ \midrule
9 & The image depicts an outdoor scene seen through a wet windshield, where water droplets obscure much of the view. This distortion leads to a significant loss of sharpness and clarity. Background elements like the car ahead and nearby structures are blurred and challenging to distinguish. Uneven focus further contributes to the poor visibility of the image. \\ \midrule
10 & This image represents an outdoor scene viewed through a wet windshield, with water droplets obscuring the visual field. The distortion caused by the droplets significantly affects sharpness and clarity, making background features like the car ahead and surrounding structures appear blurred and indistinct. The uneven focus further reduces the overall visibility of the image. \\ 
\bottomrule
\end{tabular}
\label{table:response_pool}
\end{table*}
% \clearpage

\end{document}

%% file: preamble.tex
\usepackage[dvipsnames]{xcolor}